\documentclass[journal]{IEEEtran}

\usepackage{xspace}
\makeatletter
\DeclareRobustCommand\onedot{\futurelet\@let@token\@onedot}
\def\@onedot{\ifx\@let@token.\else.\null\fi\xspace}

\def\eg{\emph{e.g}\onedot}

\makeatother
\ifCLASSINFOpdf
\else
\fi

\usepackage{amssymb}
\usepackage{amsmath}

\usepackage[colorlinks,
            linkcolor=red,
            anchorcolor=blue,
            citecolor=green]{hyperref}

\usepackage{tabu}
\usepackage{comment}
\usepackage{graphicx}
\usepackage{amsmath,bm}
\usepackage{mathtools}
\usepackage{booktabs}
\usepackage{multirow}
\usepackage{pifont}
\usepackage{color,xcolor}
\usepackage{adjustbox}
\usepackage{url}            
\usepackage{subcaption}
\usepackage{caption}
\usepackage{enumitem}
\usepackage{multicol}
\usepackage{adjustbox}
\usepackage{colortbl}
\usepackage{tikz, pgfplots}
\usepackage{cleveref}
\usepackage{makecell}

\newcommand{\Cnum}[1]{%
  \tikz[baseline=(c.base)]{
    \node[circle,draw,line width=0.35pt,inner sep=0.25ex,minimum size=1.7ex] (c) {\scriptsize #1};
  }%
}

\begin{document}
%
\title{Self-Calibrated CLIP for Training-Free Open-Vocabulary Segmentation} 
%
%
%
\author{Sule Bai, Yong Liu, Yifei Han, Haoji Zhang, \\ Yansong Tang,~\IEEEmembership{Member,~IEEE}, Jie Zhou, \IEEEmembership{Fellow,~IEEE}, and Jiwen Lu, \IEEEmembership{Fellow,~IEEE} 
	\thanks{\IEEEcompsocthanksitem The first two authors contribute equally.}
        \thanks{\IEEEcompsocthanksitem Sule Bai, Yong Liu, Yifei Han, Haoji Zhang and Yansong Tang are with the Shenzhen International Graduate School, Tsinghua University, Shenzhen, 518055, China. Email: bsl23@mails.tsinghua.edu.cn; tang.yansong@sz.tsinghua.edu.cn.}
        \thanks{\IEEEcompsocthanksitem Jie Zhou and Jiwen Lu are with the Department of Automation, Tsinghua University, Beijing, 100084, China.}
        \thanks{\IEEEcompsocthanksitem Yansong Tang is the corresponding author.}
        }

\maketitle


\begin{abstract}
Recent advancements in pre-trained vision-language models like CLIP have enabled the task of open-vocabulary segmentation. CLIP demonstrates impressive zero-shot capabilities in various downstream tasks that require holistic image understanding. 
However, due to the image-level contrastive learning and fully global feature interaction, ViT-based CLIP struggles to capture local details, resulting in poor performance in segmentation tasks.
Our analysis of ViT-based CLIP reveals that anomaly tokens emerge during the forward process, attracting disproportionate attention from normal patch tokens and thereby diminishing spatial awareness.
To address this issue, we propose Self-Calibrated CLIP (SC-CLIP), a training-free method that calibrates CLIP to generate finer representations while preserving its original generalization ability—without introducing new parameters or relying on additional backbones.
Specifically, we mitigate the negative impact of anomaly tokens from two complementary perspectives. First, we explicitly identify the anomaly tokens and replace them based on local context. Second, we reduce their influence on normal tokens by enhancing feature discriminability and attention correlation, leveraging the inherent semantic consistency within CLIP's mid-level features.
In addition, we introduce a two-pass strategy that effectively integrates multi-level features to enrich local details under the training-free setting.
Together, these strategies enhance CLIP's feature representations with improved granularity and semantic coherence.
Experimental results demonstrate the effectiveness of SC-CLIP, achieving state-of-the-art results across all datasets and surpassing previous methods by 9.5\%. Notably, SC-CLIP boosts the performance of vanilla CLIP ViT-L/14 by 6.8 times. Furthermore, we discuss our method's applicability to other vision–language models and tasks for a comprehensive evaluation. Our source code is available at \url{https://github.com/SuleBai/SC-CLIP}.
\end{abstract}



%
\IEEEpeerreviewmaketitle

\section{Introduction}
\begin{figure}[t]
    \centering
    \includegraphics[width=\linewidth]{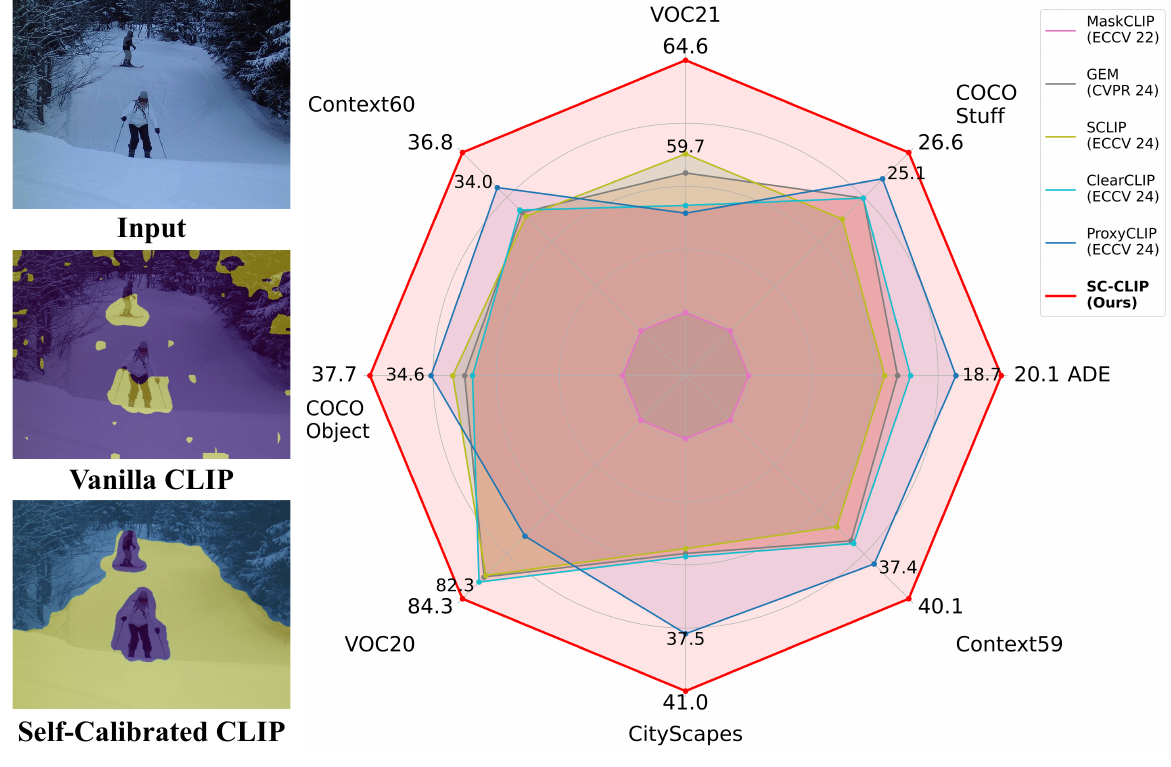}
    \vspace{-10pt}
    \caption{Left: Vanilla CLIP produces a noisy segmentation map, while our Self-Calibrated CLIP (SC-CLIP) generates a much clearer and finer result. Right: Performance comparison of the open-vocabulary segmentation methods, where our SC-CLIP achieves the best results across all benchmarks.}
    \label{fig:radar}
    \vspace{-10pt}
\end{figure}

\IEEEPARstart{O}{pen}-Vocabulary Segmentation (OVS) is an emerging task in computer vision that aims to segment arbitrary categories based on the textual inputs, overcoming the limitations of predefined category sets. To achieve this, models must generalize beyond the training data. Vision-language pretrained models such as CLIP~\cite{clip}, demonstrate remarkable zero-shot capabilities by leveraging large-scale image-text pairs, effectively fulfilling these requirements. 
However, the image-level pre-training strategy and fully global feature interactions of ViT-based CLIP lead to an excessive emphasis on global context, neglecting local and fine-grained details essential for dense prediction tasks. 
Consequently, directly applying ViT-based CLIP to segmentation tasks yields poor performance. For example, as shown in \Cref{fig:radar}, the segmentation result generated by patch-text cosine similarity exhibits considerable noise. CLIP ViT-B/16 achieves only 8.9\% mIoU on the COCO-Object dataset~\cite{caesar2018coco}, significantly lagging behind its ability on image-level recognition.

To address ViT-based CLIP's limitations in capturing local details, recent studies have proposed various modifications to its last layer. One line of research~\cite{zhou2022extract, wang2023sclip, bousselham2023grounding, li2023clip, lan2024clearclip, hajimiri2024pay, kang2024defense, shao2024explore} introduces correlative attention, replacing the original $\mathbf{Q} \mathbf{K}^\top$ attention with alternatives like $\mathbf{K} \mathbf{K}^\top$, to enhance focus on relevant regions. But these methods still operate on the global and noisy inputs, hindering their effectiveness. Another approach~\cite{lan2024proxyclip, wysoczanska2023clipdino} incorporates additional backbones~\cite{dino, sam, rombach2022high} like DINO~\cite{dino} to provide richer spatial details. Despite performance gains, they fail to fully exploit CLIP's semantic knowledge and impose extra computational costs. In fact, both strategies overlook the underlying causes of CLIP's global focus, preventing them from fundamentally resolving the issue of diminished spatial awareness in CLIP.

\begin{figure}[t]
    \centering
    \includegraphics[width=\linewidth]{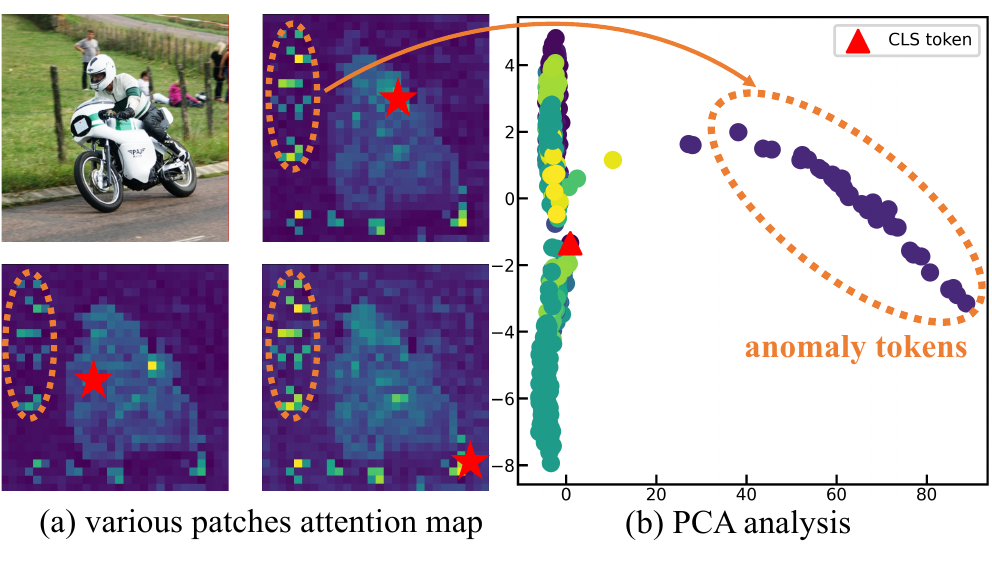}
    \vspace{-10pt}
    \caption{Anomaly tokens in CLIP. In (a), we visualize the attention maps of various selected patches (marked by \textcolor{red}{$\bigstar$}), which all exhibit excessive focus on the same regions (indicated by the \textcolor{orange}{orange dashed circle}). And this region aligns with the outliers identified in the PCA analysis shown in (b).}
    \label{fig:anomaly}
    \vspace{-5pt}
\end{figure}

Motivated by these limitations, we begin with an in-depth analysis of CLIP. As indicated by the orange dashed circles in \Cref{fig:anomaly} (a), we observe that different patch tokens consistently exhibit high activation regions within their attention map. 
These regions attract excessive attention from other normal patches, distracting their focus away from local and relevant areas.
To further investigate, we perform PCA~\cite{abdi2010principal} on the patch-level features from CLIP's last layer and project them into a 2D space, as shown in \Cref{fig:anomaly} (b), revealing these over-attended tokens significantly differ from the normal ones (including the [CLS] token). Thus, we refer to them as anomaly tokens.
We attribute CLIP's disadvantage in dense prediction tasks to the emergence of anomaly tokens, which leads to uniform attention activations across locations. This disrupts the attention's ideal ability to extract relevant semantics,  resulting in feature homogenization that diminishes local awareness and further exacerbates noise in the feature maps.

Building on this analysis, we propose enhancing CLIP's feature representation by weakening the influence of anomaly tokens. To this end, we introduce Self-Calibrated CLIP (SC-CLIP), a training-free approach that leverages CLIP's inherent properties for effective calibration, strengthening its perception on local and relevant regions.

Specifically, we address the negative impact of anomaly tokens from two perspectives. 
On the one hand, we propose directly resolving these anomaly tokens. To identify them, we apply the Local Outlier Factor (LOF) algorithm~\cite{breunig2000lof}, a method for detecting outliers. Once located, we replace these anomaly tokens with values interpolated from their spatial neighbors, considering that spatially close regions often share similar semantics. This not only serves as a regularization to prevent inappropriate attention focus but also reassigns meaningful semantic information to the anomaly tokens, aligning them with the local context.
On the other hand, to relieve the feature homogenization problem caused by anomaly tokens, we propose a self-adjusting strategy to enhance feature discriminability and attention correlation. We observe that the mid-layers' features of CLIP exhibit good semantic coherence, as shown in \Cref{fig:affinity}. In order to retain the rich semantics of deep features and the spatial consistency of mid-level features, we utilize the latter to adaptively aggregate deep features, while simultaneously enhancing attention correlation. This self-adjusting approach improves the overall semantic coherence.
Furthermore, we explore how to effectively leverage multi-level feature fusion under the training-free setting and propose a two-pass strategy to enhance the capture of details at different scales. The key insights lie in ensuring feature compatibility across layers through alignment with CLIP's final layer, and preserving the integrity of the last-layer features to maintain strong cross-modal correspondence.

Experimental results demonstrate that SC-CLIP achieves remarkable performance, establishing new state-of-the-art results across eight datasets, as shown in \Cref{fig:radar}. Our approach significantly outperforms previous methods by 9.5\% on CLIP ViT-B/16. Notably, SC-CLIP boosts the performance of vanilla CLIP ViT-L/14 by 6.8 times, without the need for additional parameters, data, or backbones.

Our contributions can be summarized as follows:
\begin{itemize}
    \item We propose SC-CLIP, a training-free method designed to enhance CLIP's dense feature representation, effectively addressing the uniform attention activations and feature homogenization caused by the anomaly tokens.
    \item We mitigate the negative effects of anomaly tokens from two perspectives. First, we explicitly address the anomaly tokens based on local context. Second, we reduce their impact on normal tokens by enhancing feature discriminability and attention correlation, leveraging the spatial consistency inherent in CLIP's mid-level features.
    \item Our approach sets new state-of-the-art results across popular benchmarks. And we conduct extensive experiments to validate the effectiveness of our method.
\end{itemize}

\begin{figure*}[t]
    \centering
    \includegraphics[width=\linewidth]{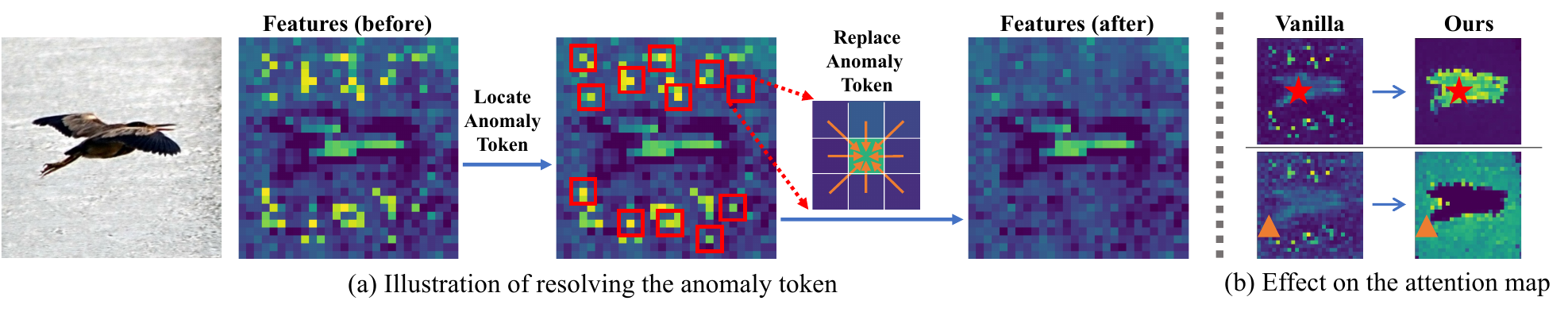}
    \vspace{-14pt}
    \caption{Resolving the Anomaly Tokens. (a) Illustration of the resolving process. We plot the feature map using the mean value of each token. After locating the anomaly tokens (the center of red square), we replace them with the interpolated values obtained from their neighboring regions. (b) Effect on the attention map. We highlight the changes on attention map for a normal token (\textcolor{red}{$\bigstar$}), and an anomaly token (\textcolor{orange}{$\blacktriangle$}).}
    \label{fig:removal}
    \vspace{-12pt}
\end{figure*}
\section{Related Work}
This section focuses on two interrelated areas: vision-language pretrained models and open-vocabulary segmentation. We highlight significant advancements and ongoing challenges, providing a critical overview that identifies gaps in current research and proposes directions for future exploration.
\vspace{-20pt}
\subsection{Vision-Language Pretrained Models}
Vision-language models~\cite{align, clip, coca, li2022blip, zhai2023sigmoid, xu2024metaclip} pretrained on large-scale web data, represented by CLIP~\cite{clip}, employ contrastive learning to align images with associated captions. These models have demonstrated remarkable zero-shot capabilities across various downstream tasks that require comprehensive image understanding, such as visual question answering and image-text retrieval~\cite{khan2022weakly, mishra2013image, antol2015vqa}. However, CLIP's image-level pre-training, which relies on the [CLS] token to represent the whole image, causes the model to excessively focus on the global features at the expense of local and fine-grained details. This limitation hinders its performance in dense prediction tasks which require pixel-level understanding. To address this, our work focuses on effectively adapting CLIP for segmentation task while preserving its original knowledge and cross-modal alignment capabilities.

\vspace{-8pt}
\subsection{Open-Vocabulary Segmentation}

The open-vocabulary segmentation (OVS) task~\cite{wu2024towards} focuses on segmenting images with arbitrary text queries by leveraging the zero-shot capabilities of vision-language models~\cite{align, clip}. This task has broad potential applications across many vision-related tasks~\cite{zhang2025occnerf, xu2024exploring, wu2024towardstip, he2024recalling, yuan2024open, liu2024universal, sun2025exploring, li2024transformer, zhao2022videoabc}. Existing works can be broadly classified into three categories: fully-supervised, weakly-supervised and training-free. Fully-supervised methods~\cite{wang2025diffusion,openseg,xu2022simple,zegformer,adapt-mask, liu2024open,san,gkc,yu2023convolutions, xie2024sed, lseg, cho2024cat, masqclip, qin2023freeseg, zhou2025rethinking, li2024omg} require fine-tuning on pixel-level annotated datasets. Weakly-supervised approaches~\cite{mukhoti2023open,ren2023viewco, tcl, groupvit, segclip, ovsegmentor, he2023clip, chen2024spatial, mukhoti2023open} reduce reliance on dense annotations by using image-text pair to guide region grouping. Training-free methods~\cite{zhou2022extract, wang2023sclip, bousselham2023grounding, li2023clip, lan2024proxyclip, wysoczanska2023clipdino, kang2024defense, shao2024explore, hajimiri2024pay, barsellotti2024training, wysoczanska2024clipdiy, lin2024tagclip, sun2024cliprnn} directly use CLIP for segmentation by making minimal adjustments to the model's architecture without additional training.

Our approach needs no training and falls into the third category. Recent methods have discovered that CLIP's final layer exhibits poor spatial consistency and proposed various modifications. For instance, SCLIP~\cite{wang2023sclip} replaces the original $\mathbf{Q} \mathbf{K}^\top$ attention with the combination of $\mathbf{Q} \mathbf{Q}^\top$ and $\mathbf{K} \mathbf{K}^\top$ attention, to enhance correlation. GEM~\cite{bousselham2023grounding} proposes generalized self-self attention and a set of regularizations. ClearCLIP~\cite{lan2024clearclip} identifies that the primary source of noise stems from the residual connections and proposes to remove it. 
CLIPTrase~\cite{shao2024explore} notices the [CLS] token may disrupt the patch correlations and proposes using self-self attention along with clustering and denoising for post-processing. 
Other works leverage additional vision models like DINO~\cite{dino} and SAM~\cite{sam} for providing fine-grained spatial details. For example, CLIP-DINOiser~\cite{wysoczanska2023clipdino} refines feature maps using the affinity learned from DINO's feature correspondence, while ProxyCLIP~\cite{lan2024proxyclip} applies them to adjust attention weights. In contrast, our approach improves semantic coherence by exploiting CLIP's internal properties and explicitly resolving the negative impact of anomaly tokens.
\section{Method}
\label{sec:method}
In this section, we begin with an overview of CLIP and its dense inference pipeline. We then present SC-CLIP, our training-free approach designed to enhance CLIP's dense representation. We address anomaly tokens from two perspectives. First, we directly identify these tokens and replace them based on local context. Next, to mitigate their influence on normal tokens, we leverage mid-level features with stronger spatial consistency to guide the adaptive aggregation of deep features and enhance the attention correlation. Furthermore, we propose a two-pass strategy to effectively integrate multi-level features and enrich spatial details.

\subsection{Preliminaries}
The CLIP ViT model encodes an input image into a token sequence \(\mathbb{X} = [x_{\text{cls}}, x_1, \dots, x_N]\), where \(x_{\text{cls}}\) is the [CLS] token and the others represent dense visual features, comprising \(N\) patch tokens. The CLIP model includes multiple layers, with each layer \(l\) processing the input \(\mathbb{X}^{(l-1)}\) as follows:
\begin{align}
   \mathbb{Z}^l &= \text{SA}(\text{LN}(\mathbb{X}^{(l-1)})) + \mathbb{X}^{(l-1)} \\
   \mathbb{X}^l &= \text{FFN}(\text{LN}(\mathbb{Z}^l)) + \mathbb{Z}^l
\end{align}
where SA, FFN, and LN denote the self-attention module, feed-forward network, and layer normalization, respectively.

For dense inference, the visual features are aligned with $C$ categories to produce the patch-text similarity map of dimensions \(N \times C\). And the final segmentation result is obtained by applying argmax operation to this similarity map.

Our method is training-free and modifies only the last layer of the CLIP visual encoder while keeping the other layers unchanged to prevent model collapse. For clarity and consistency in notation, we denote \(\mathbb{X}^{penul}\) and \(\mathbb{X}^{last}\) as the penultimate and last feature representations, respectively.

\subsection{Resolving the Anomaly Tokens}\label{sec:anomaly_removal}
We attribute CLIP's limitations in dense prediction tasks to the presence of anomaly tokens within its features, which deviate significantly from normal tokens. These anomaly tokens cause other tokens to disproportionately focus on them in deep layers, leading to identical attention activation, which undermines attention's capacity to extract semantically coherent regions. Since attention mechanism~\cite{vaswani2017attention, vit} manages spatial arrangements, this pattern intensifies noise in the feature map, ultimately degrading performance. Studies~\cite{darcet2023vision, xiao2023efficient} suggest that pretrained models like CLIP may identify redundant tokens and use them to gather global information, thereby expediting processing. However, these tokens lack semantics, conveying minimal information about their original positions. Existing methods do not explicitly address the anomaly tokens or the impact they cause to other normal tokens.

To address the negative impact of anomaly tokens, we propose an intuitive approach to directly resolve them in \(\mathbb{X}^{penul}\) before the last layer, as illustrated in \Cref{fig:removal} (a). First, it is essential to identify the anomaly tokens. As previously analyzed in \Cref{fig:anomaly}, they exhibit a clear distinction from other tokens. 
To detect these anomalies, we employ the Local Outlier Factor (LOF) algorithm~\cite{breunig2000lof}, a widely used method for anomaly detection. LOF identifies outliers by measuring the local density deviation of a data point relative to its neighbors. Specifically, it computes the local reachability density and assigns a high LOF score to points with significantly lower density than their neighbors, indicating potential anomalies. Besides, we implement the PyTorch-based LOF algorithm to enhance the computational efficiency.

Once the anomalies are located, we replace them with the values interpolated from their $3\times3$ neighboring regions, based on the assumption that features in spatially adjacent regions are generally similar. Specifically, we apply a $3\times3$ convolution kernel with the center set to 0. And if neighboring regions contain any other anomaly tokens, they are explicitly excluded from the interpolation. The operation's formula is provided below, where $\tilde{\mathbb{X}}^{penul}$ denotes the feature after operation and $\mathcal{A}$ is the set of anomaly tokens.
\begin{align}
\tilde{\mathbb{X}}_{(x, y)}^{penul} &= \frac{\sum_{i=-1}^{1} \sum_{j=-1}^{1} w_{i,j} \cdot \mathbb{X}_{(x + i, y + j)}^{penul}}{\sum_{i=-1}^{1} \sum_{j=-1}^{1} w_{i,j}}, \quad \forall (x, y) \in \mathcal{A} \notag\\
&w_{i,j} = \begin{cases} 
0, & \text{if } (x+i, y+j) \in \mathcal{A} \\
1, & \text{otherwise}
\end{cases}
\end{align}

We believe that eliminating anomaly tokens offers two key benefits. First, it acts as a form of regularization for the attention. Second, it reassigns semantic information to these anomaly tokens, aligning them with the local context. As shown in the \Cref{fig:removal} (b), the vanilla attention in CLIP has two issues: 1) normal tokens excessively focus on anomaly tokens, reducing attention on relevant local areas, and 2) anomaly tokens focus on prominent objects and other anomaly tokens. After resolving the anomaly tokens via interpolation, normal tokens refocus on relevant local regions, while the updated anomaly tokens also shift attention to appropriate areas.
\subsection{Self-Adjusting for Semantic Coherence}\label{sec:self-calibration}
After resolving the anomaly tokens, the model reduces focus on them in the last layer's attention. However, a challenge still remains: anomaly tokens have already caused substantial disruption to other normal tokens in the previous layers, diminishing their local awareness.

To alleviate this influence and further enhance feature discriminability and attention correlation, we seek to restore the spatial structure among normal tokens. Some existing methods~\cite{lan2024proxyclip, wysoczanska2023clipdino} address this by introducing backbones with strong spatial coherence, such as DINO~\cite{dino} and SAM~\cite{sam}, to provide fine-grained details. While effective, these approaches rely on additional backbones and incur extra computational costs during training and inference.

\begin{figure}[t]
    \centering
    \includegraphics[width=\linewidth]{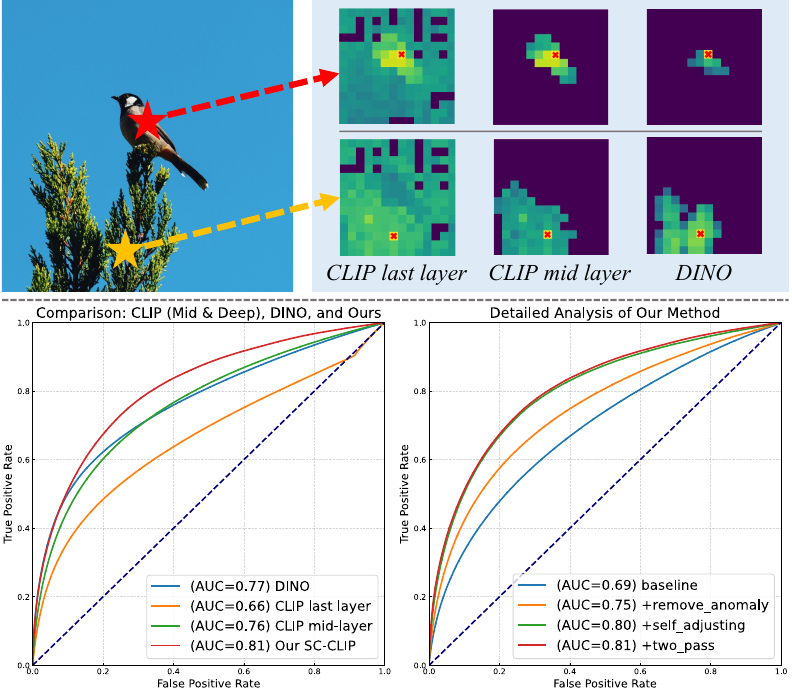}
    \vspace{-7pt}
    \caption{\underline{Top}: Visualization of patch similarities shows that CLIP's last-layer features perform poorly, but its mid-level features exhibit semantic consistency comparable to DINO. \underline{Bottom Left}: ROC curve analysis further supports this observation, with our SC-CLIP showing superior semantic coherence. \underline{Bottom Right}: Detailed ROC analysis of our method.}
    \label{fig:affinity}
\end{figure}

\begin{figure*}[t]
    \centering
    \includegraphics[width=\textwidth]{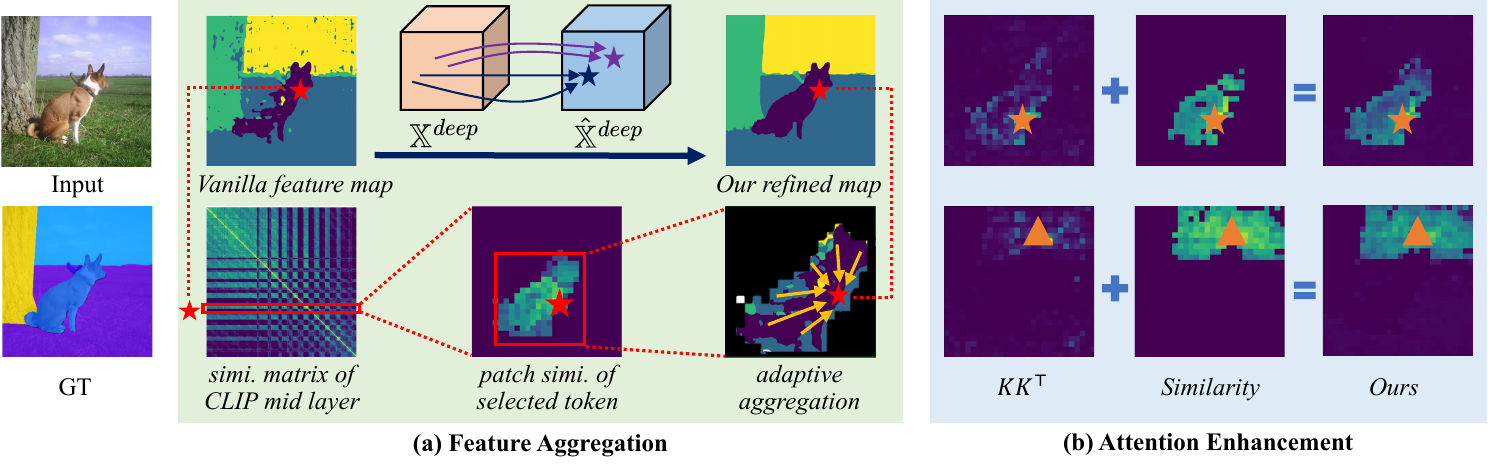}
    \vspace{-16pt}
    \caption{Illustration of the self-adjusting strategy. (a) We use the similarity map from CLIP's mid layer to adaptively aggregate deep features by combining semantically similar patches, producing clearer results. The second row provides a detailed process for the selected patch \textcolor{red}{$\bigstar$}. (b) We apply the similarity map to enhance attention, broadening and refining the activation regions.}   
    \label{fig:cali}
    \vspace{-6pt}
\end{figure*}

This motivates us to explore whether such semantic coherence can be uncovered within CLIP itself. We begin by visualizing patch similarities, as shown in \Cref{fig:affinity} (top), where we observe that CLIP's last layer performs poorly as expected. However, its mid-layer exhibits stronger semantic coherence, comparable to that of DINO. To further validate this observation, we quantitatively evaluate all features of CLIP using ROC curve analysis. Specifically, we use 2000 images from the ADE20K validation set~\cite{ade20k}. 
For each image, we extract patch-level features at each layer $l$ and compute their cosine similarity 
to construct the similarity map $\mathbf{Simi}^l\in \mathbb{R}^{N \times N}$, defined as $\mathbf{Simi}^l = \frac{\mathbb{X}^{l} \cdot \mathbb{X}^{l}}{\|\mathbb{X}^{l}\| \|\mathbb{X}^{l}\|}$.
This similarity map serves as a binary classifier to indicate whether two patches belong to the same category, with patches of the same category labeled as 1 and otherwise 0. Each patch's category is determined by majority voting on pixel labels in the segmentation map. 
A higher area under the curve (AUC) in ROC analysis reflects better semantic consistency. As shown in \Cref{fig:affinity} (bottom left), CLIP's mid-layer features exhibit strong spatial coherence (AUC=0.76), closely matching DINO (AUC=0.77). However, CLIP's last-layer features perform much worse (AUC=0.66).

The OVS task requires spatially coherent cross-modal alignment. CLIP's last feature $\mathbb{X}^{last}$ offers rich semantics but lacks coherence, while its mid-level features $\mathbb{X}^{mid}$ exhibit strong spatial consistency but are semantically limited. Inspired by~\cite{wysoczanska2023clipdino}, we propose feature aggregation to effectively combine the strengths of both. As illustrated in \Cref{fig:cali} (a), we leverage $\mathbf{Simi}^{mid}$ to adaptively aggregate $\mathbb{X}^{deep}$, generating new features $\hat{\mathbb{X}}^{deep}$ by combining semantically similar patches with weighted contributions based on their similarity. This process can be formulated as follows: 
\begin{align}
&\hat{\mathbb{X}}_{p}^{deep} = \sum_{q=1}^{N} \text{Norm}(\mathbf{Simi}_{(p,\;q)}^{mid}) \cdot \mathbb{X}_{q}^{deep}
\end{align}
The Norm function normalizes the sum to 1. The aggregated features $\hat{\mathbb{X}}^{deep}$ show better results than the original $\mathbb{X}^{deep}$.

Moreover, we find that the self-self attention (\eg, $\mathbf{K} \mathbf{K}^\top$ attention) proposed in the previous works~\cite{lan2024clearclip, wang2023sclip, li2023clip, bousselham2023grounding} exhibits insufficient attention activations. Therefore, we further incorporate the similarity map $\mathbf{Simi}^{mid}$ to augment the attention operation, as detailed in the following formula. As shown in \Cref{fig:cali} (b), the attention regions now display more extensive and accurate activations.
\begin{align}
      \text{attention\_score} = \text{softmax} ( \mathbf{K} \mathbf{K}^\top) + \text{softmax}(\mathbf{Simi}^{mid} )
   \label{eq:attn_weight}
\end{align}

Feature aggregation and attention enhancement constitute the self-adjusting strategy, which significantly improves the semantic coherence, increasing the AUC to 0.80, as shown in \Cref{fig:affinity} (bottom right). This demonstrates that the inherent consistency within CLIP's mid-level features can be effectively leveraged for adjustment.

\vspace{-2pt}
\subsection{Two-pass Strategy}\label{sec:multi}

Motivated by the effectiveness of multi-level feature aggregation in enriching details for dense prediction tasks~\cite{lin2017feature, li2024cascade, segformer, segnext, li2023clip}, we explore the potential of leveraging CLIP's multi-level features in a training-free manner.
A straightforward way is to directly sum features from different layers:
$\mathbb{X}^{last} + \sum_{i \in \mathcal{M}} \mathbb{X}^{i}$, where $\mathcal{M}$ denotes a set of mid-level features (e.g., layers 4 to 9). However, CLIP exhibits significant discrepancies across different layers. Specifically, the similarity between $\mathbb{X}^{last}$ and $\sum_{i \in \mathcal{M}} \mathbb{X}^{i}$ is merely 0.094, and directly summing them severely disrupts CLIP's cross-modal alignment capability. 
To address this issue, we note that different blocks of the ViT-based CLIP encoder capture complementary visual patterns, from local and structural cues in mid-level layers to more global and semantic information in deeper layers. The final visual layer is the only component whose output is directly contrasted with text embeddings during CLIP pre-training, and we therefore assume it effectively acts as an alignment head that maps these visual representations into the joint image–text embedding space. Based on these observations, we derive two key principles for training-free feature fusion:
\Cnum{1} Ensuring the compatibility between $\mathbb{X}^{last}$ and $\sum_{i\in\mathcal{M}} \mathbb{X}^i$ is crucial. This can be achieved by leveraging the parameter space of the last layer for alignment. Notably, the similarity between $\mathbb{X}^{last}$ and $\mathbf{L}(\sum_{i\in\mathcal{M}} \mathbb{X}^i)$ increases to 0.983, where $\mathbf{L}$ denotes the last layer. This indicates that the last layer can effectively realign the aggregated multi-level features to match the original last-layer representation, without any fine-tuning. 
\Cnum{2} Preserving the integrity of $\mathbb{X}^{last}$ is critical for maintaining cross-modal alignment, as it directly corresponds to the text embeddings.

Based on these insights, we propose a \textit{two-pass} strategy, where CLIP's final layer is explicitly employed for alignment. This strategy involves two forward passes—one with the original \(\mathbb{X}^{penul}\) and another with multi-level features, formulated as: $\mathbf{L}(\mathbb{X}^{penul}) + \mathbf{L}\left(\sum_{i \in \mathcal{M}} \mathbb{X}^{i}\right).$
Ablation experiments validate that this design enriches the representation with complementary multi-level information while preserving CLIP's cross-modal alignment capability.
\section{Experiment}

\subsection{Experimental Setup}\label{sec:settings}
\noindent \textbf{Datasets and Metric} We conduct comprehensive evaluations on eight commonly used benchmark datasets, which are grouped into two categories: 1) with a background class, including PASCAL VOC (VOC21)~\cite{pascal-voc}, PASCAL Context (Context)~\cite{pascalcontext}, and COCO Object (COCO-Obj)~\cite{caesar2018coco}; 2) without a background class, including PASCAL VOC20 (VOC20)~\cite{pascal-voc}, Cityscapes (City)~\cite{cityscapes}, PASCAL Context59 (Context59)~\cite{pascalcontext}, ADE20K (ADE)~\cite{ade20k}, and COCOStuff (COCO-Stf)~\cite{caesar2018coco}. We evaluate results using the standard \textit{mean Intersection-over-Union} (mIoU) as the metric.
\begin{table*}[t]
    \centering
    \caption{Performance comparison of our approach with other methods on eight semantic segmentation benchmarks. For a fair comparison, we reproduce all methods following the evaluation protocol in \Cref{sec:settings}, considering the different settings used by each method. CLIP-DINOiser\textsuperscript{\textdagger} denotes our reproduced results using OpenAI's pretrained weights. We report ProxyCLIP\textsuperscript{\textasteriskcentered} results using its DINO-B/16 variant.}
    \begin{tabular}{lccccccccc}
    \toprule
    \multirow{2}{*}{\textbf{Method}} & \multicolumn{3}{c}{\it With a background category} & \multicolumn{5}{c}{\it Without background category} & \multirow{2}{*}{Avg.} \\\cmidrule(lr){2-4}\cmidrule(lr){5-9}
    & VOC21 & Context & COCO-Obj & VOC20 & City & Context59 & ADE & COCO-Stf \\\midrule
    \rowcolor{lightgray} \multicolumn{10}{c}{CLIP ViT-B/16} \vspace{2pt}\\
    CLIP~\cite{clip} & 20.8 & 9.3 & 8.9 & 49.1 & 6.7 & 11.2 & 3.2 & 5.7 & 14.4 \\
    MaskCLIP~\cite{zhou2022extract} & 51.4 & 22.5 & 24.9 & 62.9 & 25.6 & 26.2 & 12.3 & 16.9 & 30.3 \\
    ReCo~\cite{reco} & 25.1 & 19.9 & 15.7 & 57.7 & 21.6 & 22.3 & 11.2 & 14.8 & 23.5 \\
    GroupViT~\cite{groupvit} & 52.3 & 18.7 & 27.5 & 79.7 & 18.5 & 23.4 & 10.4 & 15.3 & 30.7 \\
    TCL~\cite{tcl} & 51.2 & 24.3 & 30.4 & 77.5 & 23.5 & 30.3 & 14.9 & 19.6 & 33.9 \\
    CLIPSurgery~\cite{li2023clip} & 55.2 & 30.3 & 29.7 & 77.5 & 33.1 & 33.4 & 16.1 & 22.2 & 37.2 \\
    LaVG~\cite{kang2024defense} & 62.1 & 31.6 & 34.2 & 82.5 & 26.2 & 34.7 & 15.8 & 23.2 & 38.8 \\
    NACLIP~\cite{hajimiri2024pay} & 58.9 & 32.2 & 33.2 & 79.7 & 35.5 & 35.2 & 17.4 & 23.3 & 39.4 \\
    SCLIP~\cite{wang2023sclip} & 59.7 & 31.7 & 33.5 & 81.5 & 32.3 & 34.5 & 16.5 & 22.7 & 39.1 \\
    ClearCLIP~\cite{lan2024clearclip} & 57.0 & 32.2 & 32.5 & 82.3 & 32.8 & 35.8 & 17.3 & 24.0 & 39.2 \\
    GEM~\cite{bousselham2023grounding} & 58.7 & 32.0 & 32.9 & 81.7 & 32.6 & 35.6 & 16.9 & 23.9 & 39.3 \\
    CLIP-DINOiser\textsuperscript{\textdagger}~\cite{wysoczanska2023clipdino} & 57.8 & 32.1 & 32.0 & 78.0 & 36.2 & 36.0 & 16.8 & 22.6 & 38.9 \\
    ProxyCLIP\textsuperscript{\textasteriskcentered}~\cite{lan2024proxyclip} & 56.6 & 34.0 & 34.6 & 76.8 & 37.5 & 37.4 & 18.7 & 25.1 & 40.1 \\
    \rowcolor{gray!10}
    \textbf{SC-CLIP (Ours)} & \bf 64.6 & \bf 36.8 & \bf 37.7 & \bf 84.3 & \bf 41.0  & \bf 40.1 & \bf 20.1 & \bf 26.6 & \bf 43.9 \\
    \midrule
    \rowcolor{lightgray} \multicolumn{10}{c}{CLIP ViT-L/14} \vspace{2pt}  \\
    CLIP~\cite{clip} & 10.3 & 4.5 & 4.4 & 19.9 & 3.2 & 5.7 & 1.9 & 3.2 & 6.6 \\
    MaskCLIP~\cite{zhou2022extract} & 24.8 & 9.7 & 10.2 & 30.1 & 12.1 & 13.0 & 7.1 & 9.0 & 14.5 \\
    CLIPSurgery~\cite{li2023clip} & 47.9 & 27.3 & 28.1 & 84.3 & 29.7 & 31.0 & 17.3 & 21.4 & 35.9 \\
    NACLIP~\cite{hajimiri2024pay} & 52.1 & 28.7 & 29.9 & 78.6 & 31.4 & 32.1 & 17.3 & 21.4 & 36.4 \\
    SCLIP~\cite{wang2023sclip} & 44.4 & 22.3 & 24.9 & 70.6 & 21.3 & 25.2 & 10.9 & 16.5 & 29.5 \\
    ClearCLIP~\cite{lan2024clearclip} & 48.6 & 28.0 & 28.6 & 84.8 & 32.1 & 31.5 & 16.9 & 21.2 & 36.5 \\
    GEM~\cite{bousselham2023grounding} & 45.2 & 25.5 & 28.3 & 83.7 & 27.1 & 28.1 & 13.2 & 19.2 & 33.8 \\
    ProxyCLIP\textsuperscript{\textasteriskcentered}~\cite{lan2024proxyclip} & 58.1 & 34.1 & 37.4 & 82.0 & 38.1 & 37.3 & 21.2 & 25.5 & 41.7 \\
    \rowcolor{gray!10}
    \textbf{SC-CLIP (Ours)} & \bf 65.0 & \bf 36.9 & \bf 40.5 & \bf 88.3 & \bf 41.3  & \bf 40.6 & \bf 21.7 & \bf 26.9 & \bf 45.2\\
    \bottomrule
    \end{tabular}
\label{tab:main_result}
\end{table*}

\vspace{5pt}
\noindent \textbf{Implementation Details} In our experiments, we utilize CLIP~\cite{clip} with ViT-B/16 and ViT-L/14 architectures. Our code implementation is built on MMSegmentation. After deriving the similarity map $\mathbf{Simi}^{mid}$ from mid-level features, we apply thresholding following previous works~\cite{wysoczanska2023clipdino, wang2022self}. Specifically, values below the threshold $\beta$ are set to zero to strengthen feature correlation, and we set $\beta$ to 0.4. For the evaluation protocol, we adopt the sliding-window inference strategy from SCLIP~\cite{wang2023sclip}: input images are resized to have a short side of 336 (560 for Cityscapes~\cite{cityscapes} due to its higher resolution), and sliding-window inference is conducted with a $224\times224$ window and a $112\times112$ stride. No post-processing strategies are applied. Across all datasets, we use the standard ImageNet prompts~\cite{clip} combined with their category names to construct text descriptions.
We keep all hyperparameters consistent across all datasets without separate tuning.

For datasets with the background class (Pascal VOC and COCO-Object), we adhere to the evaluation protocol outlined in previous works~\cite{wang2023sclip, lan2024proxyclip, hajimiri2024pay, sun2024cliprnn}. Since the ``background'' class is overly broad for CLIP and thus challenging to classify, it is represented by a set of stuff categories such as ``sky, river, sea, ...'', which do not overlap with other target classes. And this pipeline is consistent with prior studies.
\renewcommand{\arraystretch}{1.1}
\begin{table}[t]
    \centering
    \caption{Experiments on the MESS benchmark~\cite{blumenstiel2023mess}.}
    \setlength{\tabcolsep}{4pt}
    \begin{tabular}{l|ccccc}
    \hline
        Method & SUIM & DeepCrack & Kvasir & FloodNet & Zurich \\
    \hline
        CLIP-DINOiser~\cite{wysoczanska2023clipdino} & 33.5 & 52.1 & 48.3 & 28.2 & 19.7 \\
        ProxyCLIP~\cite{lan2024proxyclip} & 35.2 & 55.0 & 52.9 & 31.0 & 21.1 \\
        \textbf{SC-CLIP (Ours)} & \textbf{36.7} & \textbf{63.1} & \textbf{57.6} & \textbf{33.4} & \textbf{22.0} \\
    \hline
    \end{tabular}
    \label{tab:mess_bench}
    \vspace{-10pt}
\end{table}

\subsection{Main Results} To ensure a fair comparison, we strictly follow the evaluation protocol specified in the \Cref{sec:settings} to reproduce all methods. For CLIP-DINOiser~\cite{wysoczanska2023clipdino}, we retrain the model using OpenAI's pretrained weights. \Cref{tab:main_result} presents the comparison of all methods, where our SC-CLIP achieves an average mIoU of 43.9\% on CLIP ViT-B/16, setting new state-of-the-art results across eight benchmarks with a notable 9.5\% improvement over previous methods. Furthermore, our method demonstrates robustness across different backbones, achieving optimal performance on CLIP ViT-L/14, improving the results by 3.5\% average mIoU.

The vanilla CLIP achieves only 14.4\% and 6.6\% mIoU on ViT-B/16 and ViT-L/14 respectively, demonstrating its limitations in capturing fine-grained spatial details. Our training-free method boosts its performance by threefold on the ViT-B/16 and by 6.8 times on the ViT-L/14.

In contrast to ProxyCLIP~\cite{lan2024proxyclip}, which relies on the DINO-B/16~\cite{dino} backbone to provide attention weights, our method does not depend on any auxiliary backbone yet achieves superior results. This highlights that the inherent properties of CLIP can be effectively leveraged to calibrate itself, thereby enhancing its feature representation.

The MESS benchmark~\cite{blumenstiel2023mess} includes a wide range of domain-specific datasets spanning fields such as earth monitoring, medical sciences, engineering, agriculture, and biology. It serves as a robust evaluation tool for assessing the generalization capability. We conduct experiments on datasets from each domain within the MESS benchmark, and as shown in \Cref{tab:mess_bench}, our SC-CLIP consistently outperforms CLIP-DINOiser~\cite{wysoczanska2023clipdino} and ProxyCLIP~\cite{lan2024proxyclip}, showing superior generalization ability.

\subsection{Ablation Study}\label{sec:ablation}
In this section, we conduct comprehensive ablation experiments to validate the effectiveness of our method. All ablations are performed on the CLIP ViT-B/16 backbone. We use SCLIP~\cite{wang2023sclip} as our baseline, which modifies the attention mechanism in the final layer by replacing the original $\mathbf{Q} \mathbf{K}^\top$ attention with $\mathbf{Q} \mathbf{Q}^\top + \mathbf{K} \mathbf{K}^\top$ attention, to enhance correlations. Additionally, we remove the residual connections and feed-forward network (FFN) following~\cite{lan2024clearclip, lan2024proxyclip}.

\renewcommand{\arraystretch}{1.1}
\begin{table}[t]
    \centering
    \caption{Ablation experiments on the proposed strategies.}
    \setlength{\tabcolsep}{8pt}
    \begin{tabular}{l|ccccc|c}
    \hline
        Method & V21 & C60 & Obj & C59 & Stf & Avg \\
        \hline
        baseline & 58.2 & 32.3 & 34.0 & 35.4 & 23.6 & 36.7 \\
        + AnomRes & 60.3 & 33.3 & 34.9 & 36.5 & 24.3 & 37.9 \\
        + AttnEnh  & 61.2  & 34.2 & 36.2 & 37.4 & 25.0 & 38.8 \\
        + FeatAgg & 62.1 & 34.8 & 37.4 & 38.0 & 25.7 & 39.6 \\
        + TwoPass & \textbf{64.6} & \textbf{36.8} & \textbf{37.7} & \textbf{40.1} & \textbf{26.6} & \textbf{41.2} \\
         \hline
    \end{tabular}
    \label{tab:ablation}
\end{table}

\renewcommand{\arraystretch}{1.2}
\begin{table}[t]
    \centering
    \caption{Ablation study on resolving the anomaly tokens.}
    \setlength{\tabcolsep}{7pt}
    \begin{tabular}{l|ccccc|c}
    \hline
        Method & V21 & C60 & Obj & C59 & Stf & Avg \\
    \hline
    \textbf{baseline} & 58.2 & 32.3 & 34.0 & 35.4 & 23.6 & 36.7 \\
    \hline
        \multicolumn{7}{c}{\cellcolor{gray!10}\textit{(a) Number of anomaly tokens}} \\
    \hline
        resolve (1) & 58.5 & 32.6 & 34.3 & 35.7 & 23.9 & 37.0 \\
        resolve (3) & 59.2 & 32.9 & 34.6 & 36.0 & 24.0 & 37.3 \\
        resolve (5) & 59.7 & 33.1 & 34.7 & 36.2 & 24.2 & 37.6 \\
        \textbf{resolve (10)} & 60.3 & 33.3 & 34.9 & 36.5 & 24.3 & \textbf{37.9} \\
        resolve (15) & 60.3 & 33.4 & 34.9 & 36.6 & 24.3 & \textbf{37.9} \\
    \hline
        \multicolumn{7}{c}{\cellcolor{gray!10}\textit{(b) Anomaly detection methods}} \\
    \hline
        Isolation Forest  & 58.9 & 32.9 & 34.5 & 36.0 & 24.1 & 37.3 \\
        DBSCAN & 58.6 & 33.1 & \textbf{35.1} & 36.1 & \textbf{24.4} & 37.5 \\
        One-class SVM  & 59.5 & 33.2 & 34.9 & 36.2 & 24.3 & 37.6 \\
        \textbf{LOF} & \textbf{60.3} & \textbf{33.3} & 34.9 & \textbf{36.5} & 24.3 & \textbf{37.9} \\
    \hline
        \multicolumn{7}{c}{\cellcolor{gray!10}\textit{(c) Interpolation methods}} \\
    \hline
        Global Mean & 56.7 & 31.9 & 33.8 & 34.8 & 23.3 & 36.1 \\
        CLS token & 59.5 & 32.2 & 34.0 & 35.9 & 23.9 & 37.1 \\
        bilinear & 59.9 & 33.2 & 34.6 & 36.3 & 24.1 & 37.6 \\
        nearest & 60.1 & 33.2 & 34.7 & 36.3 & 24.1 & 37.7 \\
        median & \textbf{60.3} & 33.2 & 34.8 & \textbf{36.5} & \textbf{24.3} & 37.8 \\
        weighted & 60.1 & \textbf{33.3} & 34.8 & 36.4 & \textbf{24.3} & 37.8 \\
        \textbf{mean} & \textbf{60.3} & \textbf{33.3} & \textbf{34.9} & \textbf{36.5} & \textbf{24.3} & \textbf{37.9} \\
    \hline
        \multicolumn{7}{c}{\cellcolor{gray!10}\textit{(d) Neighborhood sizes for interpolation}} \\
    \hline
        $\textbf{3} \times \textbf{3}$ & \textbf{60.3} & \textbf{33.3} & \textbf{34.9} & \textbf{36.5} & \textbf{24.3} & \textbf{37.9} \\
        $5 \times 5$    & 60.1 & 33.1 & \textbf{34.9} & 36.4 & \textbf{24.3} & 37.8 \\
        $7 \times 7$    & 59.8 & 33.0 & 34.8 & 36.3 & 24.1 & 37.6 \\
    \hline
    \end{tabular}
    \label{tab:combined_anomaly_study}
    \vspace{-10pt}
\end{table}

\renewcommand{\arraystretch}{1.2}
\begin{table}[t]
    \centering
    \caption{Removing anomaly tokens from different layers.}
    \setlength{\tabcolsep}{5pt}
    \resizebox{0.49\textwidth}{!}{
    \begin{tabular}{c|ccccccccccccc}
      \hline
        Methods & w/o & 1 & 2 & 3 & 4 & 5 & 6 & 7 & 8 & 9 & 10 & 11 &\textbf{12} \\
      \hline
      Self-self Attn & 20.8 & 0.3 & 1.8 & 2.5 & 0.5 & 2.6 & 2.8 & 2.6 & 1.6 & 2.3 & 13.8 & 2.0 & \textbf{58.2}\\
      \hline
          Removing Anomaly & 58.2 & 33.5 & 36.3 & 40.4 & 43.5 & 44.6 & 45.6 & 44.7 & 38.0 & 8.1 & 12.4 & 22.1 & \textbf{60.3}\\
      \hline
      \end{tabular}
    }
    \label{tab:anomaly_all_layers}
\end{table}
\renewcommand{\arraystretch}{1.3}
\begin{table}[t]
    \centering
    \caption{Ablation study of the attention enhancement.}
    \setlength{\tabcolsep}{7pt}
    \begin{tabular}{l|ccccc|c}
    \hline
        Method & V21 & C60 & Obj & C59 & Stf & Avg \\
        \hline
        \rowcolor{gray!10} \multicolumn{7}{c}{\textit{Only Correlative Self-Self Attention}} \\
        \hline
        \Cnum{1}  $\mathrm{V} \mathrm{V}^\top$ &  56.0 & 30.3 & 31.3 & 33.5 & 22.0 & 34.6 \\
        \Cnum{2} $\mathrm{Q} \mathrm{Q}^\top$ & 58.4 & 33.3 & 35.2 & 36.6 & 24.5 & 37.6 \\
        \Cnum{3} $\mathrm{Q} \mathrm{Q}^\top$+ $\mathrm{K} \mathrm{K}^\top$&  60.3 & 33.3 & 34.9 & 36.5 & 24.3 & 37.9 \\
        \Cnum{4} $\mathrm{K} \mathrm{K}^\top$ &  60.6 & 33.1 & 34.3 & 36.6 & 24.2 & 37.8 \\
        \hline
        \rowcolor{gray!10} \multicolumn{7}{c}{\textit{Only $\mathit{Simi}^{mid}$ Attention}} \\
        \hline
        \Cnum{5} $\mathrm{Simi}^{mid}$ & 60.4 & 33.8 & 35.8 & 37.1 & 24.8 & 38.3 \\
        \hline
        \rowcolor{gray!10} \multicolumn{7}{c}{\textit{Combinations of Both}} \\
        \hline
        \Cnum{1} + \Cnum{5} & 58.7 & 32.4  & 34.0 & 35.6 & 23.5 & 36.8\\
        \Cnum{2} + \Cnum{5} & 60.0 & 33.9  & 36.1 & 37.2 & 24.9 & 38.4\\
        \Cnum{3} + \Cnum{5}  & 60.9 & 33.9  & 35.9 & 37.1 & 24.7 & 38.5\\
        \textbf{\Cnum{4} + \Cnum{5}} & \textbf{61.2} & \textbf{34.2}  & \textbf{36.2} & \textbf{37.4} & \textbf{25.0} & \textbf{38.8}\\
         \hline
    \end{tabular}
    \label{tab:abla_attn}
\end{table}

\vspace{5pt}
\noindent \textbf{Analysis of Various Strategies:} In \Cref{tab:ablation}, we incrementally incorporate each strategy to highlight its contribution. First, by resolving the anomaly tokens (denoted by AnomRes), our method achieves a 1.2\% mIoU improvement, effectively addressing the issues caused by anomaly tokens, as discussed in \Cref{sec:anomaly_removal}. Next, we apply the self-adjusting strategy proposed in \Cref{sec:self-calibration}, which leverages the spatial consistency within CLIP's mid-level features to enhance attention correlation (denoted by AttnEnh), leading to a 0.9\% improvement. By adaptively aggregating deep features (FeatAgg), we obtain a further 0.8\% gain in mIoU. Finally, the two-pass strategy (TwoPass) in \Cref{sec:multi} provides a 1.6\% boost in mIoU. Collectively, these strategies contribute to a substantial 12.3\% improvement over the baseline. Besides, we conduct a detailed ROC curve analysis as shown in \Cref{fig:affinity} (bottom right). SC-CLIP achieves an AUC of 0.81, surpassing the baseline by 17.4\% and significantly enhancing the semantic coherence.

\vspace{5pt}
\noindent \textbf{Resolving the Anomaly Tokens:} To thoroughly investigate the configurations for resolving anomaly tokens, we conduct a series of ablation studies. These studies examine the impact of varying the number of anomaly tokens, the choice of different anomaly detection methods, various interpolation strategies, and different neighborhood sizes. The results are summarized in Table~\ref{tab:combined_anomaly_study}, where the first row reports the baseline results for all four groups of experiments.

(a) The number of anomaly tokens: As we use the LOF algorithm to identify the anomaly tokens, we need to adjust the contamination hyperparameter to control the number of detected anomalies. We explore the optimal number of tokens to resolve, as shown in \Cref{tab:combined_anomaly_study} (a). As the number increases from 1 to 10, performance gradually improves. However, further increasing this number to 15 yields no additional gains. Therefore, we adopt the removal of 10 tokens (about 5\% of the ViT-B/16 token sequence) in our method. 

(b) Various anomaly detection methods: We investigate alternative anomaly detection methods to the adopted LOF algorithm, including Isolation Forest, DBSCAN, and One-Class SVM. As shown in \Cref{tab:combined_anomaly_study} (b), all methods lead to improvements, with LOF algorithm achieving the best results.

(c) Various interpolation methods: We explore various interpolation methods in \Cref{tab:combined_anomaly_study} (c). After identifying the anomaly tokens, we first try replacing them with the CLS token or the mean of the normal tokens, but these strategies do not lead to performance gains. We then conduct experiments based on the local neighborhood assumption, employing bilinear, nearest-neighbor, median (using the median value of neighboring pixels), weighted (assigning weights based on distance), and mean interpolation. The results show that mean interpolation achieves the best performance, and thus we adopt it as our final choice.

(d) Different neighborhood sizes: As shown in \Cref{tab:combined_anomaly_study} (d), we compare different neighborhood sizes for mean interpolation. The results show that the $3 \times 3$ neighborhood performs better than larger ones. We hypothesize that larger neighborhoods inevitably incorporate more irrelevant regions during the interpolation process, which tends to over-smooth features and blur semantics across object boundaries.

In addition, we also test the removal of anomaly tokens starting from different layers on the VOC21 dataset in \Cref{tab:anomaly_all_layers}. The results show that the modification is only effective when removing from the last layer; otherwise, performance drops significantly. This is because the training-free setting relies on the feature modeling of earlier CLIP layers, and improvements stem from modifying only the final layer. To further support this point, we apply self-self attention~\cite{wang2023sclip, zhou2022extract, lan2024clearclip} (widely validated) to different layers. Results show that it is effective only when applied to the last layer; otherwise, it also drastically degrades performance due to feature space disruption. This suggests that the observed degradation is not caused by LOF's inability to identify anomaly tokens in other layers.

\renewcommand{\arraystretch}{1.1}
\begin{table}[t]
    \centering
    \caption{Ablation study of the feature aggregation.}
    \setlength{\tabcolsep}{7pt}
    \begin{tabular}{c|c|cccc|c}
    \hline
        Position & Source & V21 & C60 & C59 & Stf & Avg \\
        \hline
        \multicolumn{2}{c|}{without}  & 58.2  &32.3  &35.4 & 23.6 & 37.4 \\
        \hline
        \multirow{4}{*}{$\mathbb{X}^{penul}$} & \cellcolor{gray!20} DINO &  \cellcolor{gray!20}58.0 & \cellcolor{gray!20}31.8 & \cellcolor{gray!20}34.3 & \cellcolor{gray!20}23.5 & \cellcolor{gray!20}36.9 \\
        & 4 & 58.9 & 32.6  & 35.2 & 24.2 & 37.7 \\
        & \textbf{8} & 60.1  & 33.5  & 36.2 & 24.5 & 38.6 \\
        & \textbf{9} & 60.0  &33.4  &36.2 & 24.5 & 38.5 \\
        \hline
        \multirow{4}{*}{$\mathbb{X}^{last}$} & \cellcolor{gray!20} DINO  & \cellcolor{gray!20}62.6  & \cellcolor{gray!20}34.0 & \cellcolor{gray!20}37.2 & \cellcolor{gray!20} 25.4 & \cellcolor{gray!20}\textbf{39.8} \\
        & \textbf{4} & 61.4  &34.2  & 37.2 & 25.1 & 39.5 \\
        & \textbf{5} & 61.2  &34.1  & 37.1 & 25.0 & 39.4 \\
        & 9 & 60.1  &33.7  & 36.7 & 24.7 & 38.8 \\
        \hline
        \multirow{9}{*}{Both} & \cellcolor{gray!20}DINO  & \cellcolor{gray!20}58.8  & \cellcolor{gray!20}32.0  & \cellcolor{gray!20}34.5 & \cellcolor{gray!20}23.8 & \cellcolor{gray!20}37.3 \\
        & (4, 9) & 58.5  &32.5  & 35.1 & 24.1 & 37.6 \\
        & (5, 8) & 59.8  & 33.0 & 35.7 & 24.4 & 38.2\\
        & (8, 5) & 61.7  &34.2  & 37.1 & 25.1 & 39.5 \\
        & (9, 5) & 61.6 & 34.2 & 37.1 & 25.1 & 39.5\\
        & (8, 3) & 62.0 & 34.4 & 37.2 & 25.2 & 39.7 \\
        & (9, 3) & 61.9 & 34.4 & 37.2 & 25.3 & 39.7 \\
        & (8, 4) & 61.9  &34.3  & 37.1 & 25.2 & 39.6 \\
        & \textbf{(9, 4)} & 62.0  &34.4  & 37.3 & 25.3 & \textbf{39.8} \\
        \hline
    \end{tabular}
    \label{tab:adjsut_hyper}
\end{table}

\begin{figure}[t]
    \centering
    \includegraphics[width=\linewidth]{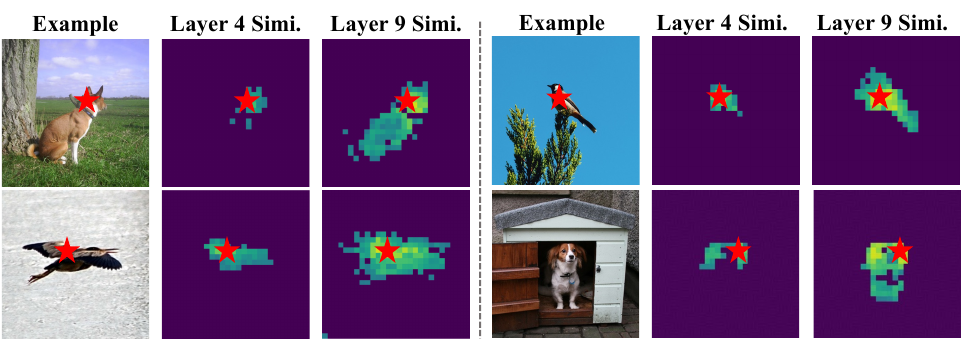}
    \vspace{-11pt}
    \caption{Visualization of shallower and deeper mid-layer patch similarity. \textcolor{red}{$\bigstar$} denotes the selected token.}
    \label{fig:supp_aff}
    \vspace{-5pt}
\end{figure}

\renewcommand{\arraystretch}{1.1}
\begin{table}[t]
    \centering
    \caption{Ablation study of the $\beta$ for feature aggregation.}
    \setlength{\tabcolsep}{4.4pt}
    \begin{tabular}{c|cccccccccc}
    \hline
        $\beta$ & w/o & 0.1 & 0.2 & 0.3 & \textbf{0.4} & 0.5 & 0.6 & 0.7 & 0.8 & 0.9 \\
    \hline
        Avg & 37.4 & 34.8 & 37.6 & 39.3 & \textbf{39.5} & 38.9 & 38.2 & 37.7 & 37.5 & 37.4 \\
    \hline
    \end{tabular}
    \label{tab:beta}
\end{table}

\renewcommand{\arraystretch}{1.1}
\begin{table}[t]
    \centering
    \caption{Ablation study of the two-pass strategy.}
    \setlength{\tabcolsep}{8pt}
    \begin{tabular}{l|ccccc|c}
    \hline
        Method & V21 & C60 & Obj & C59 & Stf & Avg \\
    \hline
        dual path~\cite{li2023clip} & 63.5 & 35.4 & 37.3 & 38.5 & 26.2 & 40.2 \\
    \hline
        \Cnum{1} baseline & 62.1 & 34.8  & 37.4 & 38.0 & 25.7 & 39.6 \\
        \Cnum{2} direct sum & 19.5 & 3.5  & 2.7  & 15.4 & 8.0 & 9.8\\
        \Cnum{3} one pass & 63.6 & 36.1 & 35.2 & 39.7 & 26.1 & 40.1 \\
        \Cnum{4} \textbf{two pass} & \textbf{64.6} & \textbf{36.8} & \textbf{37.7} &  \textbf{40.1} & \textbf{26.6} & \textbf{41.2} \\
    \hline
    \end{tabular}
    \label{tab:abla_multi}
\end{table}

\renewcommand{\arraystretch}{1.1}
\begin{table}[!htbp]
    \centering
    \caption{Selected layers of the two-pass strategy.}
    \setlength{\tabcolsep}{6pt}
    \begin{tabular}{c|cccccccccc}
    \hline
        Layers & w/o & 1-10 & 2-10 & 3-10 & \textbf{4-10} & 5-10 & 4-9 & 4-11\\
    \hline
        Avg & 42.4 & 43.8 & 43.9 & \textbf{43.9} & \textbf{43.9} & 43.8 & 43.8 & 43.6 \\
    \hline
    \end{tabular}
    \label{tab:multi_hyper}
\end{table}

\vspace{5pt}
\noindent \textbf{Self-adjusting Strategy:} To comprehensively evaluate the effectiveness of the self-adjusting strategy, we conduct ablation studies on both attention enhancement and feature aggregation.

\Cref{tab:abla_attn} presents the ablation study on attention enhancement, which can be regarded as a more detailed analysis of the ``+AttnEnh'' entry in \Cref{tab:ablation}. All experiments are conducted with anomaly tokens resolved (i.e., under the ``+AnomRes'' setting in \Cref{tab:ablation}).
We first evaluate different forms of correlative attention (\Cnum{1} to \Cnum{4} in \Cref{tab:abla_attn}), and find that these variants achieve comparable performance. We then examine using only $\mathbf{Simi}^{mid}$ (\Cnum{5}) and its integration with correlative attention. The results show that this integration (Eq.~\eqref{eq:attn_weight}) consistently outperforms either component alone, demonstrating the effectiveness of attention enhancement. Among them, enhancing $\mathbf{K} \mathbf{K}^\top$ (\Cnum{4}+\Cnum{5}) achieves the best overall performance.

And in \Cref{tab:adjsut_hyper}, we conduct a comprehensive analysis of the feature aggregation.
We find that adjusting both $\mathbb{X}^{penul}$ and $\mathbb{X}^{last}$ is essential, serving as the pre- and post-adjustment for the last layer. Experiments indicate that using deeper mid-layers (\eg, 9) for pre-adjustment and shallower mid-layer (\eg, 4) for post-adjustment yields the optimal results, corresponding to (9, 4) in the table, achieving an average 39.8\% mIoU, a 6\% improvement over the baseline. As illustrated in \Cref{fig:supp_aff}, the shallower mid-layers exhibit more localized activations, while deeper mid-layers provide broader activations. 
Based on this, we hypothesize that deeper layers are well-suited for pre-adjustment due to their broader activations, which help aggregate more regions and enable the attention mechanism to capture more coherent semantics. On the other hand, shallower layers are ideal for post-adjustment, as their localized activations can preserve intricate spatial information. 
And our method is robust to different combinations, with (8, 3) also achieving an mIoU of 39.7. 
Furthermore, we compare our approach with DINO, following prior studies~\cite{wysoczanska2023clipdino, simeoni2021localizing, wang2022self}. When applied to $\mathbb{X}^{last}$, DINO shows superior performance, showing its fine-grained details; however, our approach achieves comparable results when both pre- and post-adjustments are applied, fully unlocking CLIP's potential without relying on additional backbones.

In addition, as mentioned in \Cref{sec:settings}, we apply thresholding to $\mathbf{Simi}^{mid}$, setting values below the threshold $\beta$ to zero. To analyze the effect of the $\beta$ parameter, we experiment with different $\beta$ values using 4th-layer features to adjust $\mathbb{X}^{last}$ (row 8 of \Cref{tab:adjsut_hyper}). The results in \Cref{tab:beta} indicate that $\beta$ should not be too large, as this would cause $\mathbf{Simi}$ to activate only for itself. Conversely, if $\beta$ is too small, all tokens would be activated in $\mathbf{Simi}$, leading to feature homogenization. Overall, $\beta$ should be set to a moderate value like 0.4.

\vspace{5pt}
\noindent \textbf{Two-pass Strategy:} We conduct the ablation study of the multi-level fusion strategy in \Cref{tab:abla_multi} with the following approaches. \Cnum{1} \underline{baseline}: $\mathbf{L}(\mathbb{X}^{penul})$ \Cnum{2} \underline{direct sum}: $\mathbf{L}(\mathbb{X}^{penul}) + \sum_{i \in \mathcal{M}} \mathbb{X}^{i}$ \Cnum{3} \underline{one pass}: $\mathbf{L}(\mathbb{X}^{penul} + \sum_{i \in \mathcal{M}} \mathbb{X}^{i})$ \Cnum{4} \underline{two pass}: $\mathbf{L}(\mathbb{X}^{penul}) + \mathbf{L}\left(\sum_{i \in \mathcal{M}} \mathbb{X}^{i}\right)$. Here, $\mathbf{L}$ is the last layer, and $\mathcal{M}$ is the multi-level features.  The results align with our analysis in \Cref{sec:multi}, validating two key principles: (1) \textit{using the last layer for alignment}: comparison between \Cnum{2} and \Cnum{4} reveals that \Cnum{4} aligns the multi-level features using the parameter space of the last layer, which improves feature compatibility and leads to a significant performance boost. (2) \textit{maintaining the integrity of the last feature}: comparison between \Cnum{3} and \Cnum{4} indicates that \Cnum{3} weakens the original last feature, resulting in a performance drop.  Additionally, we compare our approach with the dual-path strategy proposed in CLIPSurgery~\cite{li2023clip}, where each layer of the image encoder in the new path is modified to self-self attention, and the outputs of the new path are directly aggregated. This can be formulated as: \underline{dual path}: $\mathbf{L}(\mathbb{X}^{penul}) + \sum_{i \in \mathcal{M}} \mathbb{X}^{i}_{new}$, where $\mathbb{X}^{i}_{new}$ is the feature from the new path. This strategy can be regarded as an enhanced version of \Cnum{2}. However, in comparison, our method achieves superior results while avoiding the additional computational overhead introduced by the new path.

\renewcommand{\arraystretch}{1.1}
\begin{table}[t]
    \centering
    \caption{Efficiency comparison of training-free methods.}
    \setlength{\tabcolsep}{7pt}
    \label{tab:Efficiency}
    \begin{tabular}{l|ccc}
    \hline
        Models & FLOPs(G) $\downarrow$  & Params(M) $\downarrow$ & Speed(FPS) $\uparrow$ \\
        \hline
        CLIP & 17.7 & 149.6 & 7.5\\
        ProxyCLIP & 34.4 & 235.4 & 3.9\\
        \hline
        \multicolumn{4}{l}{\textit{Ours}} \\ 
        \hline
        baseline & 16.8 & 149.6 & 7.8 \\
        \rowcolor{lightgray} +LOF (NumPy) & 16.8 & 149.6 & 6.3 \\
        +LOF (PyTorch) & 16.8 & 149.6 & 7.6 \\
        +AnomRes & 16.8 & 149.6 & 7.1 \\
        +SelfAdjust & 16.9 & 149.6 & 6.7 \\
        +TwoPass & 17.5 & 149.6 & 6.5 \\
    \hline
    \end{tabular}
\end{table}
\renewcommand{\arraystretch}{1.1}
\begin{table}[t]
    \centering
    \caption{Generalization to other ViT-based backbones.}
    \setlength{\tabcolsep}{7pt}
    \begin{tabular}{l|ccccc|c}
    \hline
        Method & V21 & C60 & Obj & C59 & Stf & Avg \\
    \hline
      MetaCLIP~\cite{xu2024metaclip} & 56.4 & 32.1 & 31.3 & 36.0 & 24.2 & 36.0 \\
      +Ours & 63.7 & 36.9 & 37.3 & 40.4 & 26.9 & 41.0 \\
  \hline
      OpenCLIP~\cite{cherti2023reproducible} & 55.8 & 31.4 & 31.8 & 34.8 & 23.4 & 35.4 \\
      +Ours & 60.7 & 35.2 & 36.3 & 39.2 & 27.3 & 39.7 \\

    \hline
      SigLIP~\cite{zhai2023sigmoid} & 27.1 & 11.5 & 10.9 & 18.4 & 12.0 & 16.0 \\
      +Ours & 38.3 & 19.2 & 18.0 & 24.1 & 15.9 & 23.1 \\
  \hline
      BLIP~\cite{li2022blip} & 51.1 & 28.3 & 31.2 & 29.4 & 20.2 & 32.0 \\
      +Ours & 58.7 & 34.8 & 36.7 & 37.3 & 24.9 & 38.5 \\
  \hline
    \end{tabular}
    \label{tab:generalize_backbone}
\end{table}

To analyze the impact of selected layers in multi-level fusion, we report the average mIoU across eight datasets in \Cref{tab:multi_hyper}, indicating that our method is robust to the choice of selected layers, as different combinations yield similar results. The notation 4–10 in the table denotes fusing all features from layer 4 to 10, which achieves the best 43.9\% mIoU, surpassing the baseline (w/o) by 1.5\% mIoU.


\vspace{5pt}
\noindent \textbf{Efficiency Comparison:}
As shown in \Cref{tab:Efficiency}, we compare the efficiency of different methods on the COCO-Object dataset~\cite{caesar2018coco} using an NVIDIA V100 GPU. Compared to the previous state-of-the-art method, ProxyCLIP~\cite{lan2024proxyclip}, our SC-CLIP eliminates the need for additional backbones, leading to a significant improvement in efficiency. Specifically, the FPS increases from 3.9 to 6.5, and the FLOPs are reduced from 34.4G to 17.5G. We also provide a detailed breakdown of the computational cost for each component.
By default, the LOF algorithm is implemented with NumPy, which is computationally inefficient. As discussed in \Cref{sec:anomaly_removal}, we re-implement LOF using PyTorch, resulting in a substantial inference speed-up. In particular, the FPS improves from 6.3 to 7.6, while maintaining functional consistency.

\vspace{-10pt}
\subsection{Discussion}
\begin{figure}[t]
    \centering
    \includegraphics[width=0.48\textwidth]{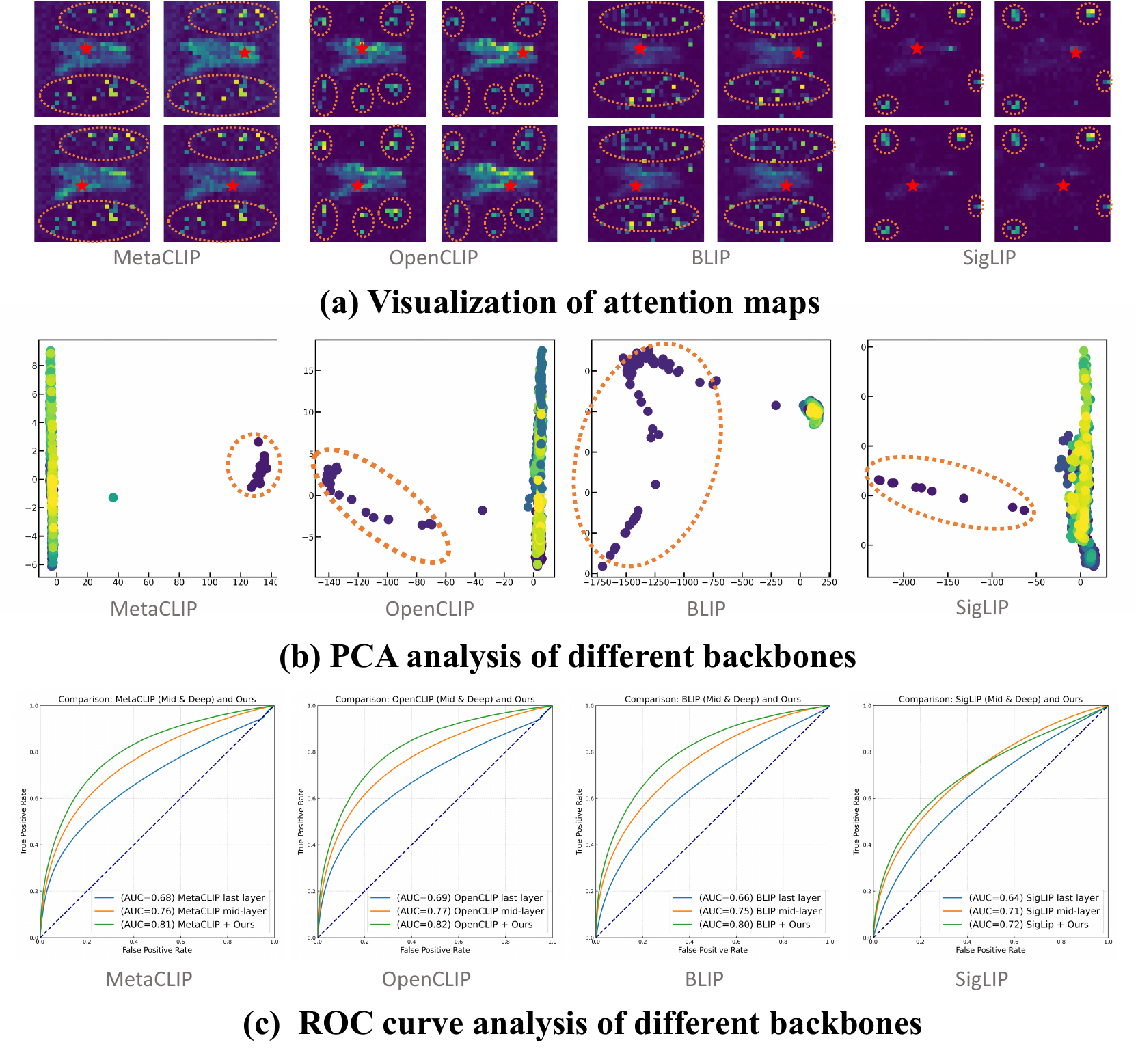}
    \caption{Detailed analysis of other ViT-based backbones.}
    \label{fig:general}
    \vspace{-3pt}
\end{figure}

\begin{figure*}[t]
    \centering
    \includegraphics[width=1\textwidth]{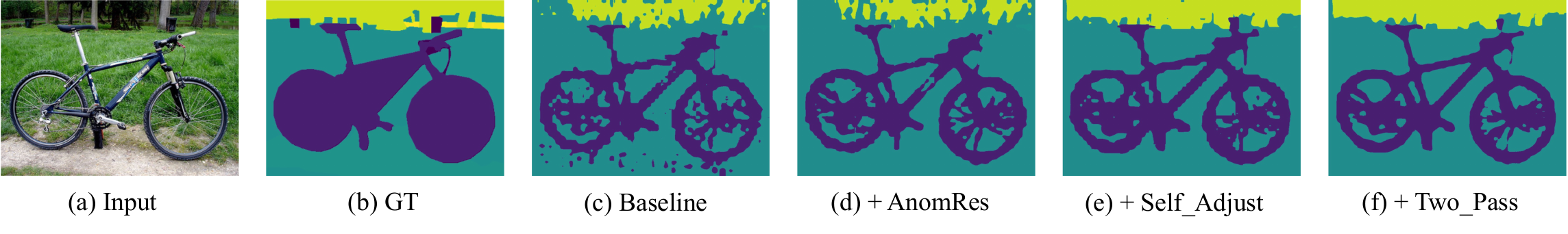}
    \vspace{-15pt}
    \caption{Qualitative results of improvement analysis, where each strategy is progressively added to demonstrate its effect.}
    \label{fig:improvement_analysis}
    \vspace{-8pt}
\end{figure*}

\begin{figure}[t]
    \centering
    \vspace{-10pt}
    \includegraphics[width=0.48\textwidth]{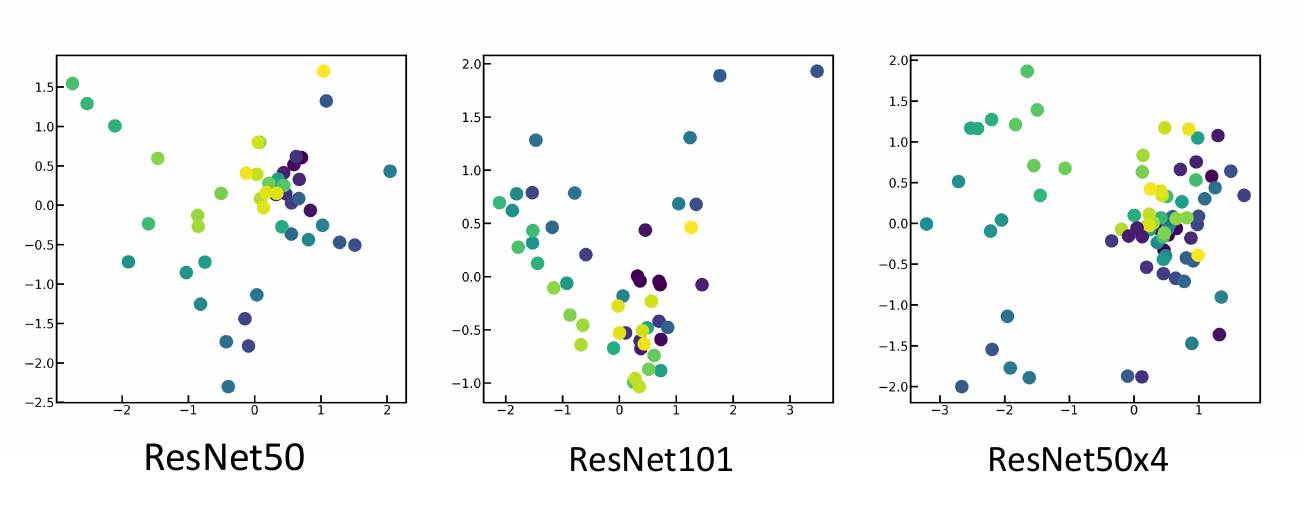}
    \vspace{-5pt}
    \caption{PCA analysis of the ResNet-based CLIP.}
    \label{fig:resnet}
    \vspace{-11pt}
\end{figure}

In this section, we provide a broader discussion on the applicability and generalization of our method. We begin by evaluating its applicability to different ViT-based vision–language models, assessing the consistency and robustness of our approach across models that vary in training data and strategies.
We then turn to ResNet-based CLIP models to examine whether the anomaly token issue also arises in non-ViT architectures. Finally, we explore the potential of our approach in open-vocabulary 3D perception tasks, demonstrating its applicability to more complex scenarios. Together, these analyses provide a more comprehensive validation for our method.


\vspace{5pt}

\noindent \textbf{Different ViT-based Vision–Language Models:} To further examine the generalization of our method, we evaluate it on several ViT-based vision–language models, which all adopt image–text contrastive pretraining but differ in certain aspects. Specifically, MetaCLIP~\cite{xu2024metaclip} and OpenCLIP (pretrained on LAION)~\cite{cherti2023reproducible} vary in the pretraining data, while BLIP~\cite{li2022blip} and SigLIP~\cite{zhai2023sigmoid} introduce alternative training paradigms or architectures. For each model, we adopt the same configurations as those used for CLIP in \Cref{sec:settings}. As shown in \Cref{tab:generalize_backbone}, our method consistently improves performance across all cases, showing its generalization to diverse models.

To further explain why our method generalizes well across different models, we provide analyses in \Cref{fig:general} (better viewed when zoomed in). In \Cref{fig:general} (a), using the image from \Cref{fig:removal} for illustration, the attention maps of four different selected patch tokens (marked by \textcolor{red}{$\bigstar$}) show that different tokens all exhibit excessive focus on the same regions, corresponding to anomaly tokens highlighted by the \textcolor{orange}{orange dashed circles}. In \Cref{fig:general} (b), PCA analysis of patch-level features reveals these anomaly tokens as clear outliers, further confirming their presence across all models. In \Cref{fig:general} (c), ROC curves comparing mid-layer and last-layer features show that mid-layer features consistently achieve higher AUC values across all models, indicating stronger discriminative ability. 
Building on this observation, our method explicitly addresses anomaly tokens and leverages mid-layer spatial consistency, leading to consistent relative AUC improvements, with gains of 19.1\% for MetaCLIP, 18.8\% for OpenCLIP, 21.2\% for BLIP, and 12.5\% for SigLIP. Overall, these results explain the root of our method's generalization ability: by tackling the consistently observed anomaly token issue and enhancing local correspondence, it achieves robust improvements across diverse ViT-based vision–language models.

\vspace{5pt}
\noindent \textbf{ResNet-based CLIP Models:} We further examine ResNet-based CLIP variants (ResNet-50, ResNet-101, and ResNet-50x4) to investigate whether the anomaly token issue also exists beyond ViT architectures. As shown in the PCA visualizations in \Cref{fig:resnet}, the feature distributions of ResNet variants are tightly clustered within a narrow range (e.g., $[-3, 3]$). In contrast, ViT-based models (\Cref{fig:general} (b)) exhibit clear outliers, with x-axis values reaching as high as 100–1000. Quantitative results in \Cref{tab:resnet} further confirm this observation: resolving anomaly tokens (denoted as ``+AnomRes'') leads to negligible changes in average performance. These findings indicate that ResNet-based CLIP does not suffer from the anomaly token issue. We hypothesize that this is due to the strong local inductive bias of convolutional architectures, which prevents the extreme outliers that often arise in ViT-based models.

\vspace{5pt}
\noindent \textbf{Open-vocabulary 3D perception tasks:} 
Open-vocabulary 3D perception has recently gained increasing attention, aiming to extend vision–language models to tasks such as 3D instance segmentation and object detection, with representative approaches including OpenMask3D~\cite{takmaz2023openmask3d}, OpenIns3D~\cite{huang2024openins3d}, Coda~\cite{cao2023coda}, and INHA~\cite{jiao2024unlocking}. To further validate the applicability of SC-CLIP in 3D scenarios, we investigate the open-vocabulary 3D semantic segmentation task using OpenScene~\cite{peng2023openscene}. In this setting, the model is required to perform per-point classification, which strongly relies on dense feature representations and naturally aligns with the design of SC-CLIP. Specifically, we employ the image feature fusion method (corresponding to the ``2D fusion'' setting in their paper), which enables a direct evaluation of the visual encoder without additional training. Experiments on the ScanNet~\cite{dai2017scannet} validation set and the Matterport3D~\cite{matterport3d} test set, as reported in \Cref{tab:scannet_matterport}, show that SC-CLIP achieves the best performance among all training-free methods. The remaining performance gap compared with training-based methods such as LSeg~\cite{lseg} and OpenSeg~\cite{openseg} can be largely attributed to their additional supervised training of both the image encoder and decoder. Overall, these results demonstrate that SC-CLIP not only advances open-vocabulary 2D segmentation, but also generalizes effectively to 3D scenarios, thereby further validating its versatility and applicability.

\begin{figure*}[t]
    \centering
    \includegraphics[width=1\textwidth]{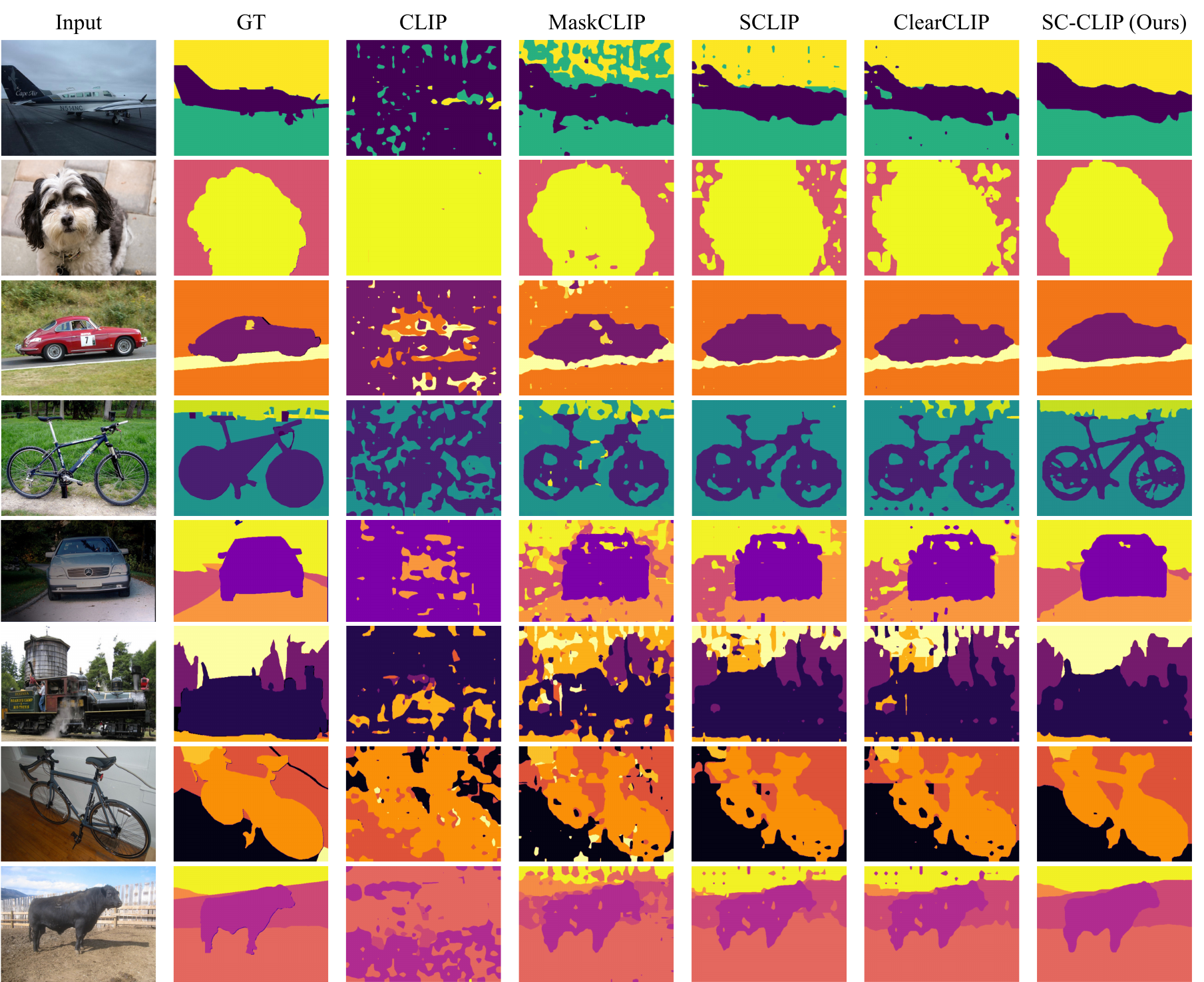}
    \vspace{-8pt}
    \caption{Qualitative Results of Open-Vocabulary Segmentation. We compare our method with CLIP~\cite{clip}, MaskCLIP~\cite{zhou2022extract}, SCLIP~\cite{wang2023sclip} and ClearCLIP~\cite{lan2024clearclip}, all without post-processing. Our SC-CLIP produces much clearer and more accurate results.}
    \label{fig:vis}
\end{figure*}

\begin{figure}[t]
    \centering
    \vspace{-2pt}
    \includegraphics[width=\linewidth]{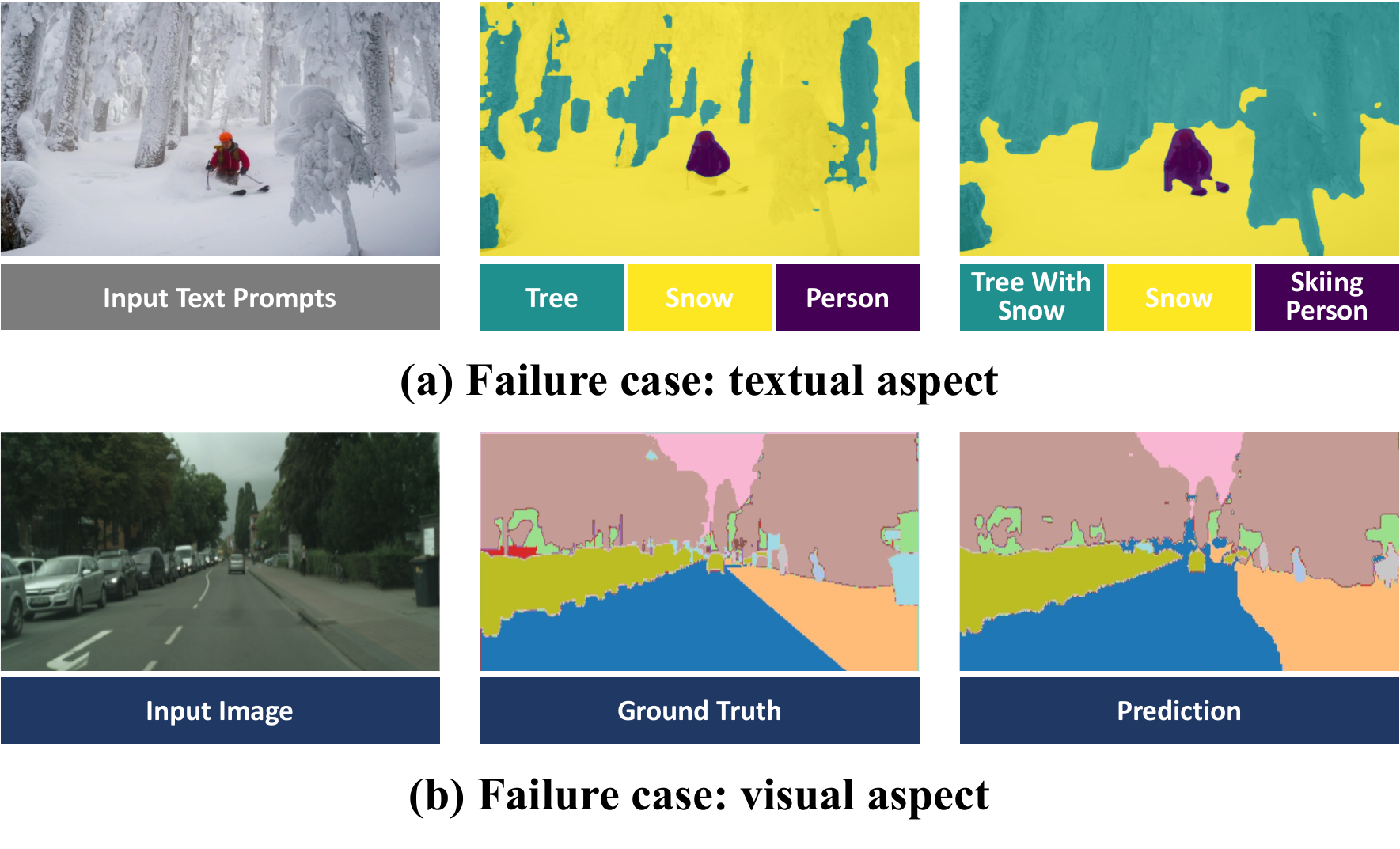}
    \vspace{-14pt}
    \caption{Failure case analysis in textual and visual aspects.}
    \vspace{-10pt}
    \label{fig:failure_case}
\end{figure}

\begin{table}[t]
\centering
\renewcommand{\arraystretch}{1.2}
\setlength{\tabcolsep}{8pt}
\caption{Resolving anomalies in ResNet-based CLIP.}

\begin{tabular}{l|ccccc|c}
\hline
Method & V20 & C60 & Obj & C59 & Stf & Avg \\
\hline
ResNet50            & 57.5 & 6.4 & 6.6 & 9.8 & 5.7 & 17.2 \\
+ AnomRes  & 57.5 & 6.5 & 6.6 & 9.8 & 5.7 & 17.2 \\
\hline
ResNet101           & 61.8 & 7.5 & 7.2 & 10.3 & 5.8 & 18.5 \\
+ AnomRes & 61.7 & 7.6 & 7.4 & 9.8  & 5.4 & 18.4 \\
\hline
ResNet50x4          & 57.6 & 6.7 & 6.3 & 9.1 & 4.9 & 16.9 \\
+ AnomRes& 57.6 & 7.1 & 6.8 & 8.8 & 4.9 & 17.0 \\
\hline
\end{tabular}

\vspace{3pt}
\label{tab:resnet}
\end{table}

\renewcommand{\arraystretch}{1.2}
\begin{table}[t]
\centering
\setlength{\tabcolsep}{9pt}
\caption{Performance comparison of different methods on open-vocabulary 3D semantic segmentation.}
\begin{tabular}{l|cc|cc}
\hline
\multirow{2}{*}{Methods} & \multicolumn{2}{c|}{ScanNet~\cite{dai2017scannet}} & \multicolumn{2}{c}{Matterport3D~\cite{matterport3d}} \\
\cline{2-5}
& mIoU & mAcc & mIoU & mAcc \\
\hline
\rowcolor{gray!30} \multicolumn{5}{c}{\textit{training-based methods}} \\
\hline
\rowcolor{gray!10}LSeg~\cite{lseg}     & 50.0 & 62.7 & 32.3 & 40.0 \\
\rowcolor{gray!10}OpenSeg~\cite{openseg}  & 41.4 & 63.6 & 32.4 & 45.0 \\
\hline
\rowcolor{gray!30} \multicolumn{5}{c}{\textit{training-free methods}} \\
\hline
CLIP~\cite{clip}      &  1.5 &  5.1 &  1.5 &  4.2 \\
SCLIP~\cite{wang2023sclip}     & 15.8 & 40.3 & 14.8 & 28.5 \\
ClearCLIP~\cite{lan2024clearclip} & 21.5 & 52.6 & 22.0 & 40.8 \\
ProxyCLIP~\cite{lan2024proxyclip} & 22.6 & 53.4 & 23.7 & 42.0 \\
SC-CLIP (Ours)      & \textbf{25.1} &\textbf{55.5} & \textbf{25.1} & \textbf{42.7} \\
\hline
\end{tabular}
\label{tab:scannet_matterport}
\vspace{-10pt}
\end{table}
\vspace{-5pt}
\subsection{Visualization}
In \Cref{fig:improvement_analysis}, we present a qualitative improvement analysis which clearly illustrates the progressive improvements introduced by each strategy and highlights the overall contributions of our method. Compared to the baseline result, resolving the anomaly tokens effectively removes the noise in the lower part of the image (d vs. c). Further applying the self-adjusting strategy enhances local correspondence and leads to better overall coherence, as can be observed in the upper region of the result (e vs. d). Finally, incorporating the two-pass strategy provides richer details, such as a more complete structure of the bicycle frame and the seat (f vs. e).

Beyond the improvement analysis, we further provide a qualitative comparison across different representative methods in \Cref{fig:vis}, including CLIP~\cite{clip}, MaskCLIP~\cite{zhou2022extract}, SCLIP~\cite{wang2023sclip}, ClearCLIP~\cite{lan2024clearclip}, and our SC-CLIP. The results demonstrate that SC-CLIP (last column) consistently produces more precise and higher-quality outcomes compared to the other methods, showing robustness across different visual contexts. In contrast, CLIP's results are considerably noisy and suffer from a homogeneity effect (\eg, in the second row, the entire image is labeled as ``dog''). Previous approaches such as SCLIP and ClearCLIP focused only on modifying the attention computation, without explicitly resolving anomaly tokens or enhancing local correspondence, which makes their results still suboptimal. By directly addressing these issues, our method achieves more semantically consistent outputs with sharper boundaries and finer structural details.

\subsection{Failure Case Analysis}
We further analyze the failure cases of our method, as illustrated in \Cref{fig:failure_case}. Our motivation is to enhance CLIP's local correspondence in a training-free manner; however, the approach is inherently bounded by CLIP's representational capacity. Specifically, the limitations manifest in two aspects: (1) \textbf{Textual aspect.} As shown in \Cref{fig:failure_case} (a), when using generic input prompts such as ``Tree'' or ``Person'', the segmentation results are suboptimal, producing vague or incomplete masks. In contrast, more specific and descriptive inputs like ``Tree With Snow'' or ``Skiing Person'' lead to significantly improved results. This phenomenon suggests that for CLIP-based models, using more specific and descriptive inputs leads to better results; (2) \textbf{Visual aspect.} The limited resolution and training paradigm of the CLIP image encoder make it difficult to accurately represent small objects or boundary regions in complex scenes. As shown in \Cref{fig:failure_case} (b), such limitations result in coarse boundaries (e.g., the contour of the car on the left) and recognition errors (e.g., failing to detect the distant traffic sign). These failure cases provide a more comprehensive perspective on the applicability and limitations of our method.

\section{Conclusion}
\label{sec:conclusion}
In this paper, we present Self-Calibrated CLIP (SC-CLIP), a training-free approach designed to enhance the open-vocabulary segmentation performance of CLIP. We observe that anomaly tokens induce uniform attention patterns and feature homogenization, which compromise spatial representation. To address this, we mitigate their adverse effects from two complementary perspectives. First, we explicitly identify anomaly tokens and replace them based on local context. Second, we reduce their influence on normal tokens by enhancing feature discriminability and attention correlation, leveraging the intrinsic semantic consistency embedded in CLIP's mid-level features. Together with a two-pass strategy, SC-CLIP achieves state-of-the-art performance without requiring additional data, parameters, or backbones. These results demonstrate that the inherent capabilities of CLIP can be effectively leveraged to calibrate itself and produce semantically coherent representations.

The motivation of this work is to enhance the dense feature representation of CLIP while preserving its original capabilities in a training-free manner. For future work, we believe that to fundamentally improve the model, modifications to the architecture and training strategy are necessary. Naturally, this would necessitate retraining the model from scratch, along with substantial computational resources and large-scale data.

\section*{Acknowledgment}
This work was supported in part by the Shenzhen Science and Technology Program under Grant CJGJZD20220517142402006, and National Natural Science Foundation of China (Grant No. 62206153).



\ifCLASSOPTIONcaptionsoff
  \newpage
\fi



%
\bibliographystyle{IEEEtran}
\bibliography{ref}
\vfill

\end{document}